\documentclass[10pt,twocolumn,letterpaper]{article}
\usepackage[pagenumbers]{cvpr} % To force page numbers, e.g. for an arXiv version

\usepackage{graphicx}
\usepackage{amsmath}
\usepackage{amssymb}
\usepackage{booktabs}
\usepackage{times}
\usepackage{float}

\usepackage{microtype}
\usepackage{epsfig}

\usepackage{algorithm}
\usepackage{algorithmic}

\usepackage{bm}
\usepackage{multirow}
\usepackage{xcolor}

\newcommand{\G}{\bm{G}}
\newcommand{\D}{\bm{D}}
\newcommand{\E}{\bm{E}}
\newcommand{\w}{\bm{w}}
\newcommand{\x}{\bm{x}}

\newcommand{\FID}{\textbf{FID$\downarrow$}}  % Notation of FID score in Table.
\newcommand{\SWD}{\textbf{SWD$\downarrow$}}  % Notation of MSE score in Table.

\usepackage[pagebackref,breaklinks,colorlinks]{hyperref}

\usepackage[capitalize]{cleveref}
\crefname{section}{Sec.}{Secs.}
\Crefname{section}{Section}{Sections}
\Crefname{table}{Table}{Tables}
\crefname{table}{Tab.}{Tabs.}

\begin{document}

%\title{Clothing Model Generation from Unpaired Data for E-shops}
\title{Weakly Supervised High-Fidelity Clothing Model Generation}

 \author{Ruili Feng$^1$,Cheng Ma$^2$,Chengji Shen$^2$,Xin Gao$^3$,Zhenjiang Liu$^3$,\\
 Xiaobo Li$^3$,Kairi Ou$^3$,Zhengjun Zha$^1$\\
 $^1$University of Science and Technology of China, $^2$Zhejiang University, $^3$Alibaba Group \\
 ruilifengustc@gmail.com,\{cheng.ma,chengji.shen\}@zju.edu.cn, \\ \{zimu.gx,stan.lzj,xiaobo.lixb\}@alibaba-inc.com, mailokr@gmail.com, zhazj@ustc.edu.cn
 }
% mailokr@gmail.com,zhaodeli@gmail.com,zhazj@ustc.edu.cn}

% \email{ruilifengustc@gmail.com,{cheng.ma,chengji.shen}@zju.edu.cn,\\
% {zimu.gx,stan.lzj,xiaobo.lixb}@alibaba-inc.com,\\
% mailokr@gmail.com,zhaodeli@gmail.com,zhazj@ustc.edu.cn}

%\author{Ruili Feng\\
%University of Science and Technology of China\\
%Anhui, China\\
%{\tt\small ruilifengustc@gmail.com}
%\and
%Cheng Ma, Chengji Shen\\
%Zhejiang University\\
%Hangzhou, China\\
%{\tt\small \{cheng.ma,chengji.shen\}@zju.edu.cn}
%\and
%Xin Gao, Zhenjiang Liu, Xiaobo Li, Kairi Ou\\
%Alibaba Group\\
%Hangzhou, China\\
%{\tt\small \{zimu.gx,stan.lzj,xiaobo.lixb\}@alibaba-inc.co%m, mailokr@gmail.com}
%\and
%Zhengjun Zha\\
%University of Science and Technology of China\\
%Anhui, China\\
%{\tt\small zhazj@ustc.edu.cn}
%}
\maketitle
% \twocolumn[{
% \renewcommand\twocolumn[1][]{#1}
% \maketitle
% \thispagestyle{empty}
% \begin{center}
%   \centering
%   \includegraphics[width=1\linewidth]{figure1_v4.pdf}\vspace{-0.4cm}
%   \captionof{figure}{Given a set of commercial models with underwear and different clothing images, the proposed method can generate realistic clothing model results with clear pattern reconstruction.}\label{fig:teaser}
% \end{center}
% }]

%%%%%%%%% ABSTRACT
\begin{abstract}
  The development of online economics arouses the demand of generating images of models on product clothes, to display new clothes and promote sales. However, the expensive  proprietary model images challenge the existing image virtual try-on methods in this scenario, as most of them need to be trained on considerable amounts of model images accompanied with paired clothes images. In this paper, we propose a cheap yet scalable weakly-supervised method called Deep Generative Projection (DGP) to address this specific scenario. Lying in the heart of the proposed method is to imitate the process of human predicting the wearing effect, which is an unsupervised imagination based on life experience rather than computation rules learned from supervisions. Here a pretrained StyleGAN is used to capture the practical experience of wearing. Experiments show that projecting the rough alignment of clothing and body onto the StyleGAN space can yield photo-realistic wearing results. Experiments on real scene proprietary model images demonstrate the superiority of DGP over several state-of-the-art supervised methods when generating clothing model images.
\end{abstract}

\begin{figure}[t]
  \centering
  \includegraphics[width=1\linewidth]{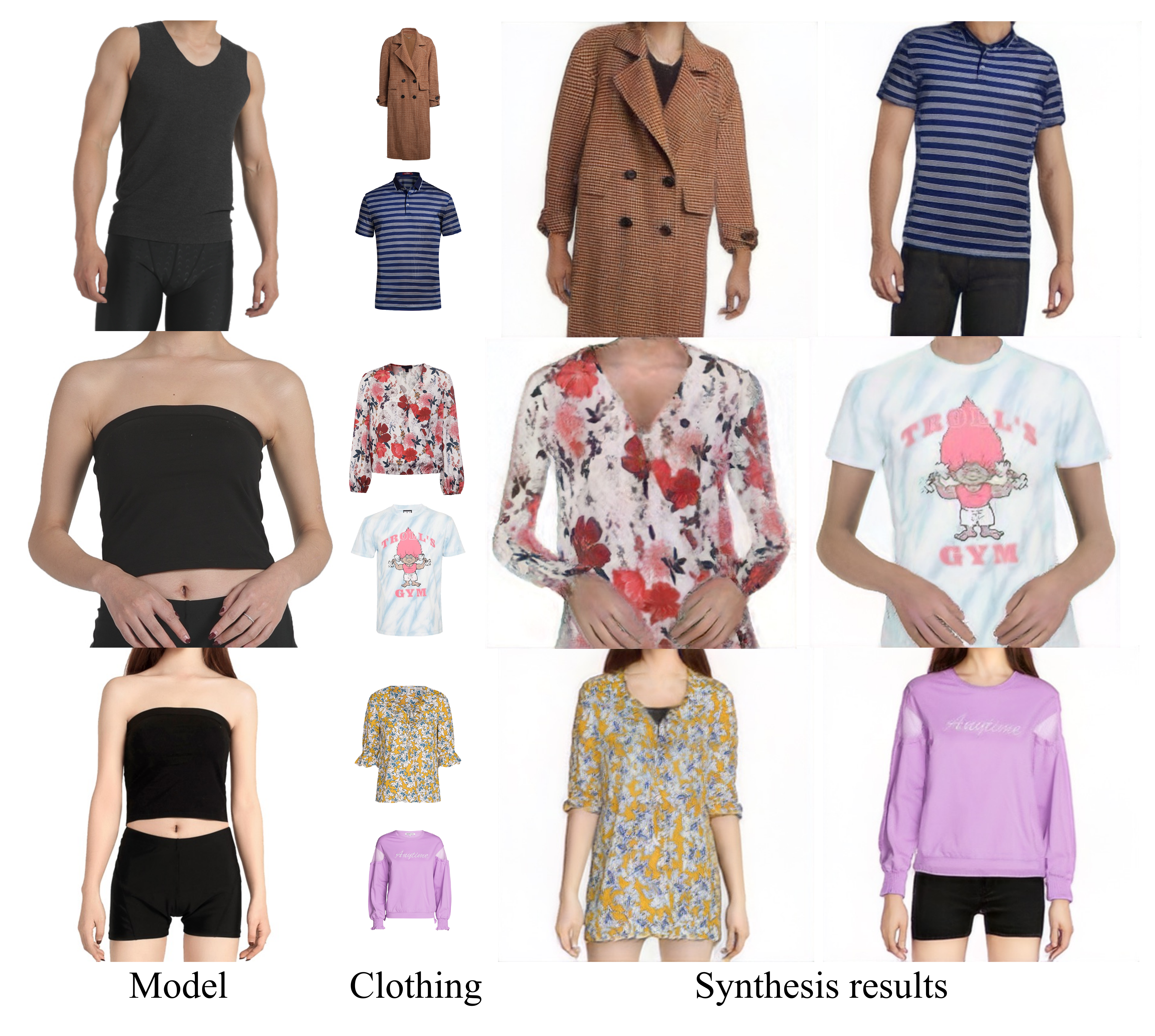}
  \caption{Given a set of commercial models with underwear and different clothing images, the proposed method can generate realistic clothing model results with clear pattern reconstruction.}\label{fig:teaser}
\end{figure}

%%%%%%%%% BODY TEXT
\section{Introduction}

\label{sec:intro}
Taking pictures of models on product clothes is an essential demand for online apparel retails to display clothes and promote sales. However, it is highly expensive to hire models and professional studios to actually take those pictures of wearing each clothing. Thus image virtual try-on \cite{han2018viton,wang2018toward,santesteban2019learning,dong2019towards,hauswiesner2013virtual,neuberger2020image,jae2019viton,yang2020towards,dong2019fw,gao2021shape}, the technique that generates wearing results from clothing and model images, has rapidly aroused broad academic and industrial interests.

A practical algorithm, however, must consider the cost to apply it in  industrial scenarios. Existing image virtual try-on (VTO) methods are highly expensive to train. Most of them \cite{han2018viton,yang2020towards,minar2020cp,jae2019viton,wang2018toward,issenhuth2020not} are trained on paired image data of clothing and a person on that clothing. Such paired data consume considerable labor cost thus is infeasible to collect at scale. Also, the testing data of clothing model generation should be proprietary model images as only they can be legally used in the final display on e-shop websites. Those images are highly expensive due to hiring commercial models and buying all necessary rights. Thus algorithms should avoid relying on those proprietary model images during training, while managing to bear the non-negligible performance drop incurred by discrepancy in testing and training.

In response to those challenges, we propose Deep Generative Projection (DGP), a powerful weakly-supervised method to yield realistic try-on results while training on cheap unpaired data collected from the Web, which is motivated by the procedure of people predicting how they will look like while picking clothes. It is an imagination based on life experience rather than rules learned from paired annotations. People may pick up the clothes and align the clothes to their shoulders or necks, from which they then imagine the picture of wearing those clothes. Following this idea, the DGP method reproduces this process in the clothing model generation scenario. Given a clothing image and a proprietary model image, we first align the clothing to the model's body dominated by a simple perspective transformation \cite{mezirow1978perspective} of four body key points. Then we project this rough alignment onto the synthesis space of a pretrained StyleGAN \cite{karras2019style,karras2020analyzing}. The StyleGAN is trained on abundant unsupervised fashion images collected from the Web. Thus, it represents real-world knowledge of wearing. A couple of semantic and pattern searches will then yield the realistic clothing model image, as is shown in Fig. \ref{fig:teaser}. The whole algorithm needs no paired data or proprietary model images during training, thus it is practical for the industrial scenario.

In conclusion, the contributions of this paper include:
\begin{itemize}
    \item We propose the first framework to generate clothing model images for online clothing shops, which has not received enough attention in the virtual try-on community;
    \item The proposed method consumes only unpaired data, and no proprietary model images during the training, which is more practical for industrial applications than most existing methods;
    \item Our weakly-supervised method significantly outperforms some state-of-the-art supervised competitors in both numerical and visual quality, and demonstrates good robustness under preprocessing mistakes.
\end{itemize}

\section{Related Works}
\paragraph{Virtual Try-on}
Virtual try-on methods can be broadly divided into 3D-based methods \cite{guan2012drape,patel2020tailornet,brouet2012design,pons2017clothcap,patel2020tailornet,rohmer2010animation} and 2D image-based \cite{wang2018toward,han2018viton,choi2021viton,jae2019viton,yu2019vtnfp,yang2020towards,dong2019towards,han2019clothflow} methods. As 3D methods often induce extra resources in collecting data or physical simulation, 2D methods are generally more popular. Many existing 2D methods \cite{han2018viton,jae2019viton,ge2021parser,choi2021viton,dong2019fw,han2019clothflow} split the try-on procedure into a warping procedure and a synthesis procedure. The warping procedure learns to deform the target garment to fit the figure of the model image, while the synthesis procedure tries to merge the warped garment image with the model image. Such methodology demands paired data \cite{han2018viton} to supervise the training of warping modules. While recent advances in GAN image synthesis also inspire 2D methods based on pretrained GANs. VOGUE \cite{lewis2021tryongan,lewis2021vogue} adopts interpolation to search a latent code that can generate the target clothing in the latent space of a pretrained StyleGAN. StylePoseGAN \cite{sarkar2021style} and pose with style \cite{albahar2021pose} explore the rich style space of pretrained StyleGANs to manipulate the poses of synthesis images. Those works often omit the discussion of encoder or inversion technique to send the original images to StyleGANs. They also often have trouble in precisely reconstructing the pattern of clothes, and lose certain semantic information of the person or target clothing images. 

\paragraph{StyleGAN} StyleGAN \cite{karras2019style,karras2018progressive,karras2020analyzing} is the dominating method in unconditional image synthesis. Since it is proposed, broad interests have been attracted to use StyleGANs in various domains of image manipulations \cite{tewari2020stylerig,wu2021stylespace,shen2020interpreting,yang2020towards}. Most previous works find that the style space \cite{shen2020interpreting,tewari2020stylerig}, a feature layer after the first 8-layer MLP of the StyleGAN generator, reveals fascinating semantic controlling over synthesis images. Subsequent studies \cite{wu2021stylespace,richardson2021encoding} also confirm deeper layers of StyleGAN generator owning similar or even stronger ability. As the preprocessing of manipulating real images by StyleGAN, inverting real images to the style space of StyleGAN also receives special attention. Image2StyleGAN \cite{abdal2019image2stylegan,abdal2020image2stylegan++} finds the inverted style code through solving an optimization problem based on distance metrics. While pSp \cite{richardson2021encoding} and e4e \cite{tov2021designing} train explicit encoders to obtain the style code, and claim that explicit encoders can acquire more meaningful semantics for subsequent manipulations.
\begin{figure*}[ht]
    \centering
    \includegraphics[width=1\linewidth]{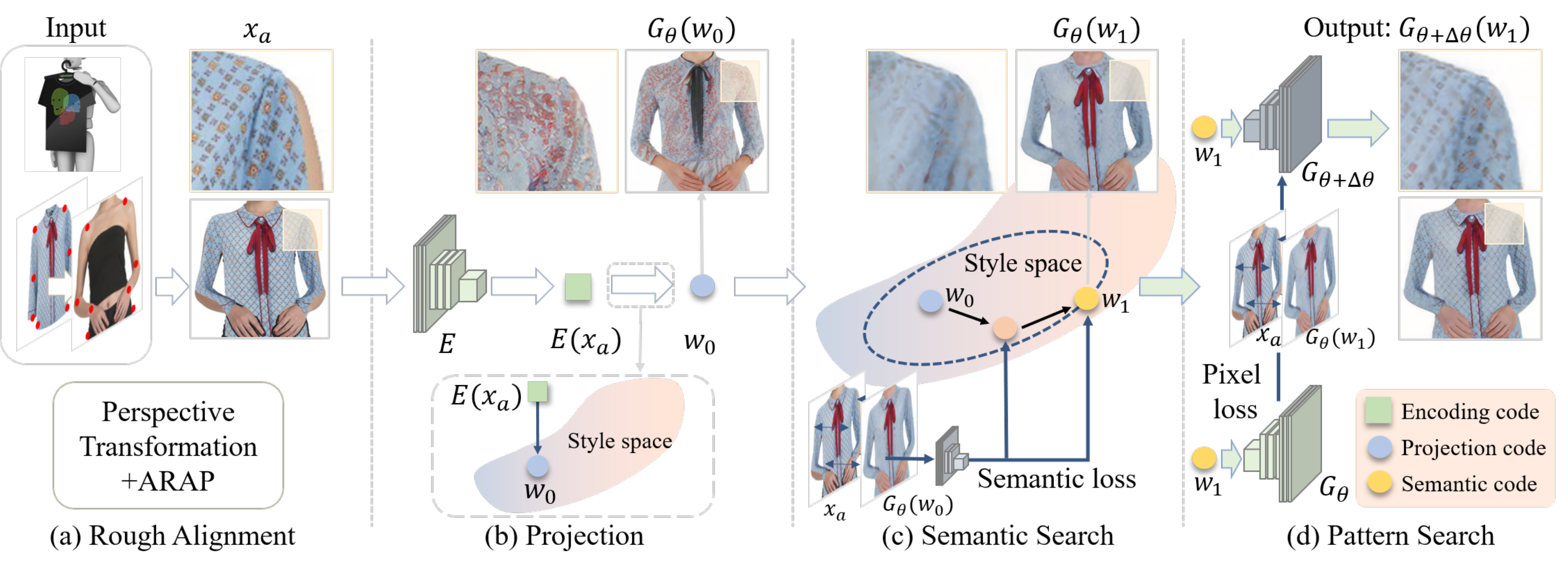}%\vspace{-0.2cm}
    \caption{Framework of the proposed DGP method. A rough alignment $\bm{x}_a$ (a) of model and clothing images is fed into a novel projection operator (b), which truncates flaws of the aligned image, and transfers it into a projection code $\bm{w}_0$ that yields realistic synthesis and similar semantics on the StyleGAN synthesis space $\bm{G}_{\bm{\theta}}$. This process is implemented by projecting the encoding code $\bm{E}(\bm{x}_a)$ of a pretrained encoder $\bm{E}$ onto the high-density region of style space. A semantic search (c) then solves a constraint optimization problem on the synthesis space of StyleGAN to find the semantic code $\w_1$ that recovers missing semantics. A pattern search (d) further adjusts parts of the StyleGAN parameters from $\bm{\theta}$ to $\bm{\theta}+\Delta\theta$. The new synthesis space $\bm{G}_{\bm{\theta}+\Delta\theta}$ then precisely reconstructs patterns of the original clothing in $\bm{G}_{\bm{\theta}+\Delta\theta}(\w_1)$. $\bm{G}_{\bm{\theta}+\Delta\theta}(\w_1)$ is the final output of the DGP method.}
    \label{fig:frame}%\vspace{-0.15cm}
\end{figure*}

%-------------------------------------------------------------------------

\section{Task Setting}\label{sec:task}
In this paper, the new proposed clothing model generation task is different from typical VTO scenarios. Differences are elaborated as follows.
% The clothing model generation we tackle in this paper is different from typical VTO scenarios. Thus it is good to discuss and strengthen those differences before we start.
\paragraph{Gaps Between Training and Testing Environments} The proprietary model images are too costly to construct a sufficiently large training set. Thus algorithms should avoid relying on those images during training. However, they need to test performance on those images, as only proprietary model images can be legally used in the product display. Generally, the solution to the clothing model generation task should be stable under such a dilemma.

\paragraph{Original Clothes of Models} Models wearing thick long sleeve clothes can perturb the image generation process. While the task here is irrelevant to the original clothes of models, we only consider cases on simple sleeveless clothes like underwear or vest.

\paragraph{Benchmark} The Commercial Model Image dataset (CMI) is collected to serve as a benchmark for real scene applications. The CMI dataset includes 2,348 images of models on underwear, including different genders, ages, body shapes, and poses. All model images are taken in professional studios and granted portrait rights.
In addition, we also collect 1,881 clothing images with clean backgrounds from e-commerce platforms, evenly containing 16 categories, and corresponding category annotations. There are no paired relations between the clothing images and model images, and both those images are unavailable during the training phase. Please consult the supplementary materials for details of this dataset.

\section{Deep Generative Projection}
\paragraph{Overview} Given a model image and a clothing image, we follow the procedure that people predict try-on results, and decompose it as a fast first impression, and a further mulling over the impression. To imitate the first impression, a novel projection operator is employed on the StyleGAN space. It projects a rough alignment of clothing and body onto the synthesis space of a pretrained StyleGAN. Different from typical literature in encoding GANs, here we do not pursue an exact reconstruction of the rough alignment, but a domain that preserves similar semantics yet maintains synthesis fidelity. The `warping' of the proposed method is actually accomplished in this step, as realistic synthesis always yields realistic wearing. The further mulling of impression is conducted by two fine-grained information searches in the neighborhood of the encoder projection. One is the semantic search in the feature space of the StyleGAN. It recovers semantic information lost in the projection phase. The other is the pattern search in the parameter space of the StyleGAN to reconstruct the pattern of clothing. By strictly constraining these two steps in the neighborhood of the encoder projection, we can precisely reconstruct the target clothing while preserving the fidelity of the usual StyleGAN synthesis. A simple review of the proposed method is provided in Fig. \ref{fig:frame}.

% \begin{figure}
%     \centering
%     \includegraphics[width=1\linewidth]{rough align.png}\vspace{-0.2cm}
%     \caption{The rough alignment step here is an imitation of the preprocessing of human try-on, where people usually align the clothes to the key points of their bodies.}
%     \label{fig:rough align}\vspace{-0.15cm}
% \end{figure}

\paragraph{Rough Alignment} The rough alignment aligns the clothing and model at key points of neck, hip, elbow, and wrist, as shown in Fig. \ref{fig:frame} (a). The alignment of the neck and hip key points is implemented by a perspective transformation \cite{mezirow1978perspective}, while the alignment of elbow and writs is implemented by the As Rigid As Possible (ARAP) \cite{alexa2000rigid,igarashi2005rigid,jacobson2011bounded} algorithm. The ARAP is a classical non-parametric deformation algorithm, which is efficient in controlling key point alignment. For different types of clothes, the alignment rule admits slight differences. For example, sleeveless clothes do not involve alignment on the elbow and wrist. If hands or arms are in front of the body, they will further be cropped and stuck on top of the aligned image to maintain consistency. See the supplementary materials for more details.

\paragraph{Training of StyleGAN} To train the StyleGAN, we collect an E-Shop Fashion (ESF) dataset of 180,000 clothing model images from the Internet. The images are all cropped to the region between jaw and thigh, and resized to the resolution of $512\times512$. The whole dataset is split into 170,000 training samples and 10,000 testing samples. The StyleGAN is trained on the training dataset and the training terminates at the FID score of 2.16. More details about the StyleGAN and ESF dataset can be found in the supplementary materials.

\subsection{Projection}\label{sec:projection}
The projection is the key to the success of the DGP method, as it offers a compact and rich domain for the subsequent mulling of details. This section gives a theoretical perspective of good projection in our task.

\paragraph{High-density Region of Style Space} Following previous studies on StyleGAN synthesis \cite{shen2020interpreting,tewari2020stylerig,abdal2019image2stylegan}, we focus on projecting images to the style space $\mathcal{W}+$ of the StyleGAN generator $\bm{G}$. The style space $\mathcal{W}+$ is the feature layer produced by the first 8-layer MLP of the StyleGAN generator. It reveals fascinating disentanglement of semantic features \cite{shen2020interpreting,karras2020analyzing}. Different from GAN inversion \cite{richardson2021encoding,abdal2019image2stylegan,abdal2020image2stylegan++, tov2021designing} techniques focusing on the exact reconstruction of input, here we care more about synthesis fidelity instead of reconstruction accuracy. This is because the exact reconstruction of rough alignment is useless for our task. As a feature space, points in the style space $\mathcal{W}+$ are not uniformly distributed. Previous works \cite{shen2020interpreting,abdal2019image2stylegan,karras2020analyzing} have demonstrated that regions of higher sampling probability density can yield much more plausible synthesis than those of lower density. To strike a good balance between fidelity and similarity, the projection should always land on the high-density region of style space $\mathcal{W}+$, and give reasonable strength to each semantic component. 

Thus, instead of projecting the rough alignment onto $\mathcal{W}+$ directly, we propose to project it onto each of the principal components of $\mathcal{W}+$ space. The projection on each component is further truncated if it far exceeds the average strength of $\mathcal{W}+$ space on that component. We will prove later how this operation can help anchor the projection inside the high-density region. 

\newcommand{\q}{\bm{q}}

\paragraph{The Projector} Rigorously, we sample five million points on the $\mathcal{W}+$ space of the StyleGAN, and compute the PCA decomposition \cite{wold1987principal} of those points. We then get the mean value $\bm{\mu}$ of $\mathcal{W}+$, covariance matrix $\bm{\Sigma}$, and a set of principal components $\bm{Q}=(\q_1,...,\q_n)$ together with their strengths stored in $\bm{\Lambda}=diag\{\sigma_1,...,\sigma_n\}$, where $n$ denotes the dimension of $\mathcal{W}+$, and $\bm{\Sigma}=\bm{Q}\bm{\Lambda}\bm{Q}^T$. Then, instead of training an encoder $\bm{E}$ to learn the style code directly, we propose to learn a series of principal strengths $\bm{s}=(s_1,s_2,...,s_n)^T$ of a given image, and truncate those principal strengths to reproduce the style code in an appropriate region. Given a rough alignment image $\bm{x}_a$, the style code $\w_0$ then can be computed as%\vspace{-0.1cm}
\begin{gather}
    \bm{s}=\bm{E}(\x_a),\\
     \w_0=Tr(\bm{q_1}s_1\sqrt{\sigma_1}+...\bm{q}_n s_n\sqrt{\sigma_n})+\bm{\mu}\\
     =\bm{Q}\bm{\Lambda}^{\frac{1}{2}}Tr(\bm{s})+\bm{\mu},%\vspace{-0.1cm}
\end{gather}
where $Tr$ is a truncation operator with cutoff coefficient $\psi>0$, such that%\vspace{-0.1cm}
\begin{align}
Tr(\bm{v})=
\left\{\begin{array}{ll}
     \bm{v}, \Vert \bm{v}\Vert_2<\psi,\\
     \frac{\bm{v}}{\Vert\bm{v}\Vert_2}\psi, \Vert \bm{v}\Vert_2\geq\psi.
\end{array}\right.%\vspace{-0.1cm}
\end{align}
For simplicity, we will call the operation%\vspace{-0.1cm}
\begin{equation}
    \w=\bm{P}(\x)=Tr(\bm{Q}\bm{E}(\x_a))+\bm{\mu}%\vspace{-0.2cm}
\end{equation}
as projection, and $\bm{P}$ the projector. Given a rough alignment image $\bm{x}_a$, the projector $\bm{P}$ is used to project it onto the synthesis space of StyleGAN as %\vspace{-0.2cm}
\begin{equation}
    \w = \bm{P}(\x_a).%\vspace{-0.4cm}
\end{equation}

\paragraph{Property of the Projector} The projector may lose certain information represented by the small principal components after truncation, but it will enforce the projection landing on the high-density region of the style space. Rigorously, we have the following theorem:
\newtheorem{theorem}{Theorem}
\begin{theorem}
Assume that $\mathcal{W}+$ follows the multi-variable Gaussian distribution, then the output of the projector $\bm{P}$ will always fall in the high-density region of $\mathcal{W}+$, which is an $n$-dimensional ellipse $\mathcal{E}$ with axes $\q_1,...,\q_n$, and axis lengths $\psi \sigma_1^{-\frac{1}{2}},...,\psi \sigma_n^{-\frac{1}{2}}$. Rigorously, let $\omega_{n-1}$ denote the volume of the $n-1$ dimensional unit ball, for a random sample $\w$ from $\mathcal{W}+$, the possibility of it outside $\mathcal{E}$ is%\vspace{-0.2cm}
\begin{equation}
\begin{aligned}
    \mathbb{P}(\w \notin \mathcal{E})
    =\mathbb{P}(\chi_n^2>\psi^2),%\vspace{-0.3cm}
\end{aligned}
\end{equation}
where $\chi_n^2$ is the n-dimensional Chi-square distribution \cite{lancaster2005chi}, and $\mathbb{P}(\chi_n^2>\psi^2)$ drops to zero drastically as $\psi$ grows larger; and for an arbitrary input $\bm{x}$, we have%\vspace{-0.2cm}
\begin{equation}
    \bm{P}(\bm{x})\in\mathcal{E}=\{\w:(\w-\bm{\mu})^T\bm{\Sigma}^{-1}(\w-\bm{\mu})\leq\psi^2\}.%\vspace{-0.2cm}
\end{equation}
\end{theorem}

\paragraph{Training of the Projector} We employ a simple ResNet50 \cite{he2016deep} architecture for the encoder network $\bm{E}$, and train the projector $\bm{P}$ on the training data distribution $p_{data}$ of the pretrained StyleGAN $\bm{G}$. The training loss consists of a pixel similarity $\mathcal{L}_p$, a perceptual similarity $\mathcal{L}_f$, an attribute similarity $\mathcal{L}_{attr}$, and an adversarial fidelity $\mathcal{L}_{adv}$:%\vspace{-0.2cm}
\begin{gather}
     \min_{\bm{P}}  \lambda_p \mathcal{L}_{p}+\lambda_f\mathcal{L}_f + \lambda_{attr}\mathcal{L}_{attr}+ \lambda_{adv}\mathcal{L}_{adv},%\vspace{-0.3cm}
\end{gather}
where $\lambda_p,\lambda_f,\lambda_{attr},\lambda_{adv}$ are hyperparameters.
The pixel similarity is directly captured by $l_2$ distance in the pixel space:%\vspace{-0.1cm}
\begin{equation}
    \mathcal{L}_p=\mathbb{E}_{\x\sim p_{data}}[\Vert\G(\bm{P}(\x))-\x\Vert_2^2].%\vspace{-0.1cm}
\end{equation}
The perceptual similarity is captured by a pretrained VGG16 network $\bm{V}$. The VGG16 \cite{simonyan2014very} network $\bm{V}$ is trained on ImageNet \cite{deng2009imagenet}, and the final convolution layer is taken as the feature space to compute the similarity:%\vspace{-0.1cm}
\begin{align}
    \mathcal{L}_f=\mathbb{E}_{\x\sim p_{data}}[\Vert\bm{V}(\G(\bm{P}(\x)))-\bm{V}(\x)\Vert_2^2].%\vspace{-0.1cm}
\end{align}
The attribute similarity is captured by a pretrained clothing attribute classifier $\bm{R}$, which is trained on the FashionAI \cite{zou2019fashionai} dataset. It is a simple ResNet50 \cite{he2016deep} architecture that identifies seven different attributes of clothing, like types of the sleeve and neckline. The final convolution layer is taken as the feature space to compute the similarity: (see supplementary materials for the detail of the attribute classifier)%\vspace{-0.1cm}
\begin{equation}
    \mathcal{L}_{attr}=\mathbb{E}_{\x\sim p_{data}}[\Vert\bm{R}(\G(\E(\x)))-\bm{R}(\x)\Vert_2^2].%\vspace{-0.1cm}
\end{equation}
The adversarial fidelity loss is computed through carrying on the adversarial game of pretrained StyleGAN generator $\bm{G}$ and discriminator $\bm{D}$. The discriminator $\bm{D}$ is asked to distinguish the projected images $\bm{G}(\bm{P}(\bm{x}))$ from real images, and the projector to try to fool the discriminator by projecting images to regions of higher fidelity. The generator network $\bm{G}$ is frozen during training, while $\bm{P}$ and $\bm{D}$ are alternatively optimized as in training a typical GAN \cite{goodfellow2014generative}:%\vspace{-0.2cm}
\begin{equation}
    \begin{aligned}
       \mathcal{L}_{adv}=\max_{\D}\mathbb{E}_{\x\sim p_{data}}&\left[\log(1-\D(\G(\bm{P}(\x))))\right.\\
    &+\left.\log(\D(\x))\right].
\end{aligned}%\vspace{-0.2cm}
\end{equation}

\begin{figure}[t]
  \centering
  \includegraphics[width=1\linewidth]{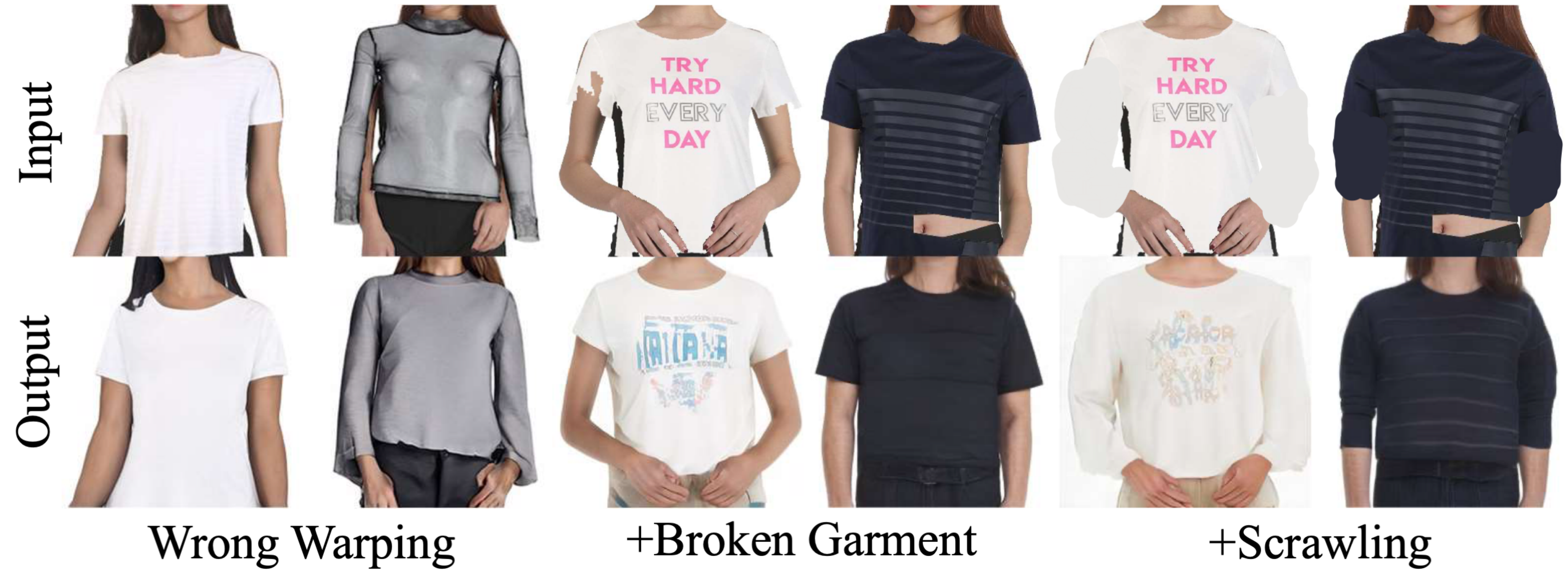}%\vspace{-0.2cm}
  \caption{The imagination ability of the projector. The first row shows the inputs, each of which has some unrealistic defects. The second row shows the results of the projection. Although the encoder fails to preserve all semantics and details of the original images, it always generates plausible outputs thanks to it staying inside the high-density region of StyleGAN. }\label{fig: encoder-imagination}%\vspace{-0.15cm}
\end{figure}

\paragraph{Imagination Ability of the Projector} The projector can serve as a fascinating feature extractor of arbitrary image input. Though it is only trained to reconstruct real images, it can also extract semantic features from unreal images produced by splicing, scrawling, or warping. The extracted features can then be sent to the generator to reproduce those semantics in a plausible way. Fig. \ref{fig: encoder-imagination} illustrates this ability of the projector. This ability is endowed by forcing the output of the encoder to stay inside the high-density domain of the generator during training. Thus whatever inputs will be projected to a suitable domain of the generator knowledge that only produces plausible images.

%%%%%%%%%%%%%%%%%%%%%%%%%%%%%%%%%%%%%%%%%%%%%%%%

%%%%%%%%%%%%%%%%%%%%%%%%%%%%%%%%%%%%%%%%%%%%%%%

\begin{figure}[t]
  \centering
  \includegraphics[width=1\linewidth]{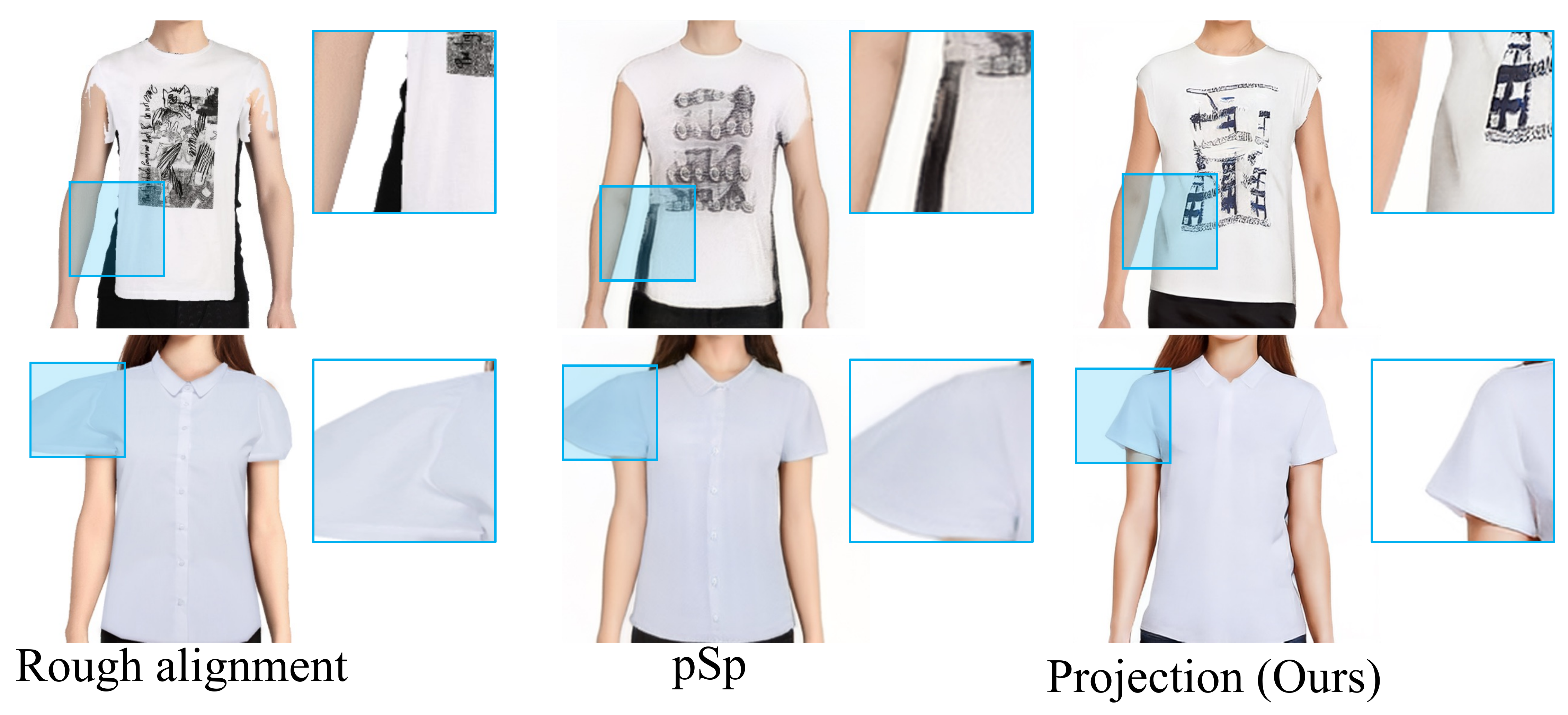}%\vspace{-0.2cm}
  \caption{Projector vs. pSp \cite{richardson2021encoding}. pSp will generate a less plausible but more similar reconstruction to the input image. While the projector always generates plausible images regardless of the inputs.}\label{fig: figure_encoder_structure}%\vspace{-0.15cm}
\end{figure}

\paragraph{Projector vs. the SOTA Encoder} As we have explained, the projector is designed to encourage fidelity rather than accuracy of reconstruction. Here we compare its reconstruction with the stat-of-the-art StyleGAN encoder pSp \cite{richardson2021encoding}. The results are reported in Fig. \ref{fig: figure_encoder_structure}. When handling unrealistic rough alignment results, the projector produces far more plausible results than pSp. On the other hand, pSp is faithful to the rough alignment, thus can inherit the unrealistic effects and generate images of low fidelity.

\subsection{Semantic Search}
The style code $\w$ discovered by the projector can only reproduce some high-level semantics (such as the style and category of the garment, pose of the model). To obtain fine-grained semantics, we need an optimization-guided search inside the neighborhood of the projection. The optimization problem is %\vspace{-0.25cm}
\begin{gather}\label{eq:semantic-search}
    \min_{\w\in \mathcal{C}}\eta_p l_p+\eta_f l_f+\eta_{attr} l_{attr}+\eta_{adv} l_{adv},\\
    l_p=\Vert W*\G(\w)-W*\x_a\Vert_2^2,\\
    l_f=\Vert \bm{V}(W*\G(\w))-\bm{V}(W*\x_a)\Vert_2^2,\\
    l_{attr}= \Vert \bm{R}(W*\G(\w))-\bm{R}(W*\x_a)\Vert_2^2,\\
    l_{adv}=\log[1-\bm{D}(\G(\w))],%\vspace{-0.25cm}
\end{gather}
where $\mathcal{C}$ is the neighborhood of $\w$, $\bm{R}$ and $\bm{V}$ are pretrained clothing attribute classifier and VGG16 network introduced in the Sec. \ref{sec:projection} respectively, $\x_a$ is the rough alignment of clothing and model, $\eta_p,\eta_f, \eta_{attr}, \eta_{adv}$ are hyperparameters, and $W$ is a dynamic spatial weight matrix to adjust the strength of optimization among different regions. Central regions of the body and clothing have higher weights, while marginal regions have tiny weights. This design allows the generator to adjust the marginal contents of synthesis according to central regions. We enforce the limitation of $\w$ staying in $\mathcal{C}$ to maintain the whole optimization inside the StyleGAN domain of high synthesis fidelity.

\paragraph{Dynamic Spatial Weight $W$} The dynamic spatial weight matrix is an exponential function over the intersection of the model body and the aligned clothing. Let $I$ denote the identity function for intersection of body and clothing, $\partial I$ denote its boundary, and $d((i,j),\partial I)$ denote the distance of pixel position $(i,j)$ to the boundary of $I$, then $W$ is computed as %\vspace{-0.15cm}
\begin{gather}
    W_{ij}=\left\{\begin{array}{ll}
     1-\exp(-d((i,j),\partial I)^2),& I(ij)=1, \\
     0,& I(ij)=0.
\end{array}\right.%\vspace{-0.15cm}
\end{gather}
More detailed instructions on computing $W$ can be found in the supplementary materials.

%%%%%%%%%%%%%%%%%%%%%%%%%%%%%%%%%%%%%%%%%%%%%%%%

\begin{figure}[t]
  \centering
  \includegraphics[width=1\linewidth]{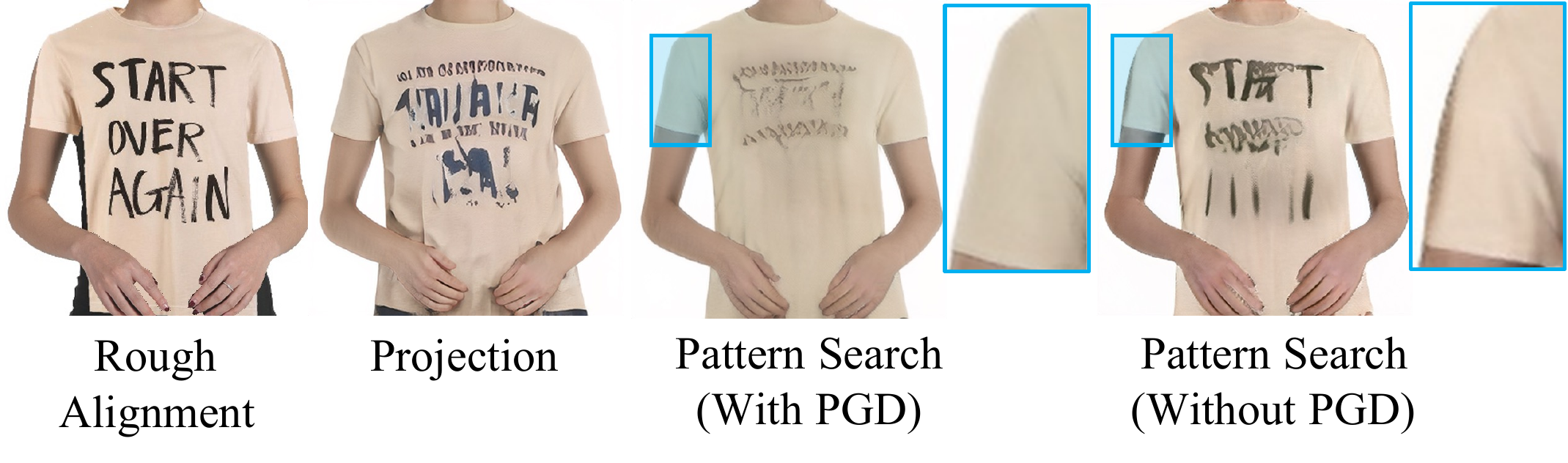}%\vspace{-0.3cm}
  \caption{Influence of the Projected Gradient Descent (PGD) \cite{madry2017towards}. Without PGD, the result of pattern search will tend to overfit the original rough alignment image, which is what we do not want in this phase, as the warping of rough alignment is usually wrong.}\label{fig: PGD-ablation}%\vspace{-0.15cm}
\end{figure}

%%%%%%%%%%%%%%%%%%%%%%%%%%%%%%%%%%%%%%%%%%%%%%%

\paragraph{Solve the Constrain $\mathcal{C}$}To make sure that the semantic search stays in the high-density region of StyleGAN knowledge space, we use a constraint optimization strategy that has been widely used in the field of adversarial attacks \cite{guo2019simple, dong2018boosting,madry2017towards}. Given a convex optimization problem \cite{boyd2004convex}%\vspace{-0.1cm}
\begin{gather}\label{eq:pgd}
    \min_{\bm{w}}f(\bm{w})\,\, s.t.\,\,\bm{w}\in\mathcal{C},%\vspace{-0.4cm}
\end{gather}
where $\mathcal{C}$ is a convex set, we can solve it by projecting the updated parameter onto $\mathcal{C}$ after each iteration of gradient descent, as shown in Algorithm \ref{algorithm} \cite{madry2017towards}.
%%%%%%%%%%%%%%%%%%%%%%%%%%%%%%%%%%%%%%%%%%%%%%%
\begin{algorithm}[t]
	\caption{Projected Gradient Descent.}
	\label{algorithm}
	\begin{algorithmic}
		\STATE {\bfseries Input:} Hyperparameter $\gamma$, objective $f(\bm{w})$, convex constraint region $\mathcal{C}$, initial point $\bm{w}_0\in\mathcal{C}$, counter $k=0$.%\vspace{-0.2cm}
		\REPEAT
		\STATE Compute the gradient of $f$ at $\bm{w}_k$ as $\nabla f(\bm{w}_k)$.
		\STATE Update $\bm{w}_k$ by%\vspace{-0.3cm}
		\begin{equation}
		    \bm{w}_{k+1}=\bm{w}_k-\gamma\nabla f(\bm{w}_k).%\vspace{-0.5cm}
		\end{equation}
		\STATE Project $\bm{w}_{k+1}$ back to $\mathcal{C}$ by %\vspace{-0.3cm}
		\begin{equation}\label{eq:projection}
		    \bm{w}_{k+1}=\arg\min_{\bm{w}\in\mathcal{C}}\Vert \bm{w}_{k+1}-\bm{w}\Vert.%\vspace{-0.5cm}
		\end{equation}
		\STATE Update counter as $k = k+1$.
		\UNTIL{Convergence.}
		\STATE {\bfseries Output:} The numerical solution $\bm{w}_k$ of problem (\ref{eq:pgd}).
	\end{algorithmic}
\end{algorithm}
%%%%%%%%%%%%%%%%%%%%%%%%%%%%%%%%%%%%%%%%%%%%%%%%

In the experiments, we find that the spherical neighborhood is good enough to constrain the optimization inside the high-density region of StyleGAN knowledge space. Thus we set $\mathcal{C}$ as a ball $B(\w_0,4)$ centered at the projector output $\w_0$ with radius 4. Note that problem (\ref{eq:projection}) then has a closed-form solution: %\vspace{-0.2cm}
\begin{align}
&\w_{k+1}=\arg\min_{\w\in\mathcal{C}}\Vert\w_{k+1}-\w\Vert\\
&=
\left\{\begin{array}{ll}
     \w_0+4\frac{\w_{k+1}-\w}{\Vert \w_{k+1}-\w\Vert_2},& \Vert \w_{k+1}-\w_m\Vert_2 > 4, \\
     \w_{k+1},& \Vert \w_{k+1}-\w\Vert_2\leq 4.
\end{array}\right.%\vspace{-0.2cm}
\end{align}
Problem (\ref{eq:semantic-search}) then can be solved efficiently by Algorithm \ref{algorithm}.

\paragraph{Necessity of Neighborhood Constraint $\w\in\mathcal{C}$} The neighborhood constraint $\w\in\mathcal{C}$ is very important in the semantic search. Fig. \ref{fig: PGD-ablation} reports the results with and without it. Without this constraint and the Projected Gradient Descent (PGD), the optimization quickly runs out of the high-density region of the generator and produces implausible details.

\subsection{Pattern Search}
The pretrained generator contains rich semantic information. However, for a specific pattern like characters, we may not be able to reconstruct it accurately by semantic search. Our strategy here turns to optimizing some key parameters $\bm{\theta}$ of the generator to `overfit' such pattern. We find that optimizing the side-way noise injection parameters \cite{karras2019style,abdal2019image2stylegan,abdal2020image2stylegan++} in the StyleGAN Network is a good choice. Those parameters are proven to decide the local details and stochastic variations of generated images, as is carefully studied in the original paper of StyleGAN \cite{karras2020analyzing,karras2019style}. As our purpose is to `overfit' the pattern, here we do not need a feature or attribute loss to close the semantics. The optimization only involves pixel loss and adversarial loss to preserve fidelity:%\vspace{-0.15cm}
\begin{equation}
    \begin{aligned}
        \min_{\bm{\theta}\in B(\bm{\theta}_0,4)}\eta_p\Vert W*\G_{\bm{\theta}}(\w)-W*\x_a\Vert_2\\
        +\log(1-\D(\G_{\bm{\theta}}(\w))),
    \end{aligned}%\vspace{-0.15cm}
\end{equation}
where $\bm{\theta}_0$ is the initial value of those parameters in the pretrained StyleGAN, and $B(\bm{\theta}_0,4)$ is a ball centered at it with radius 4. We again solve the problem with Algorithm \ref{algorithm}. The result is the final output of the DGP pipeline. 

\begin{figure}[t]
  \centering
  \includegraphics[width=1\linewidth]{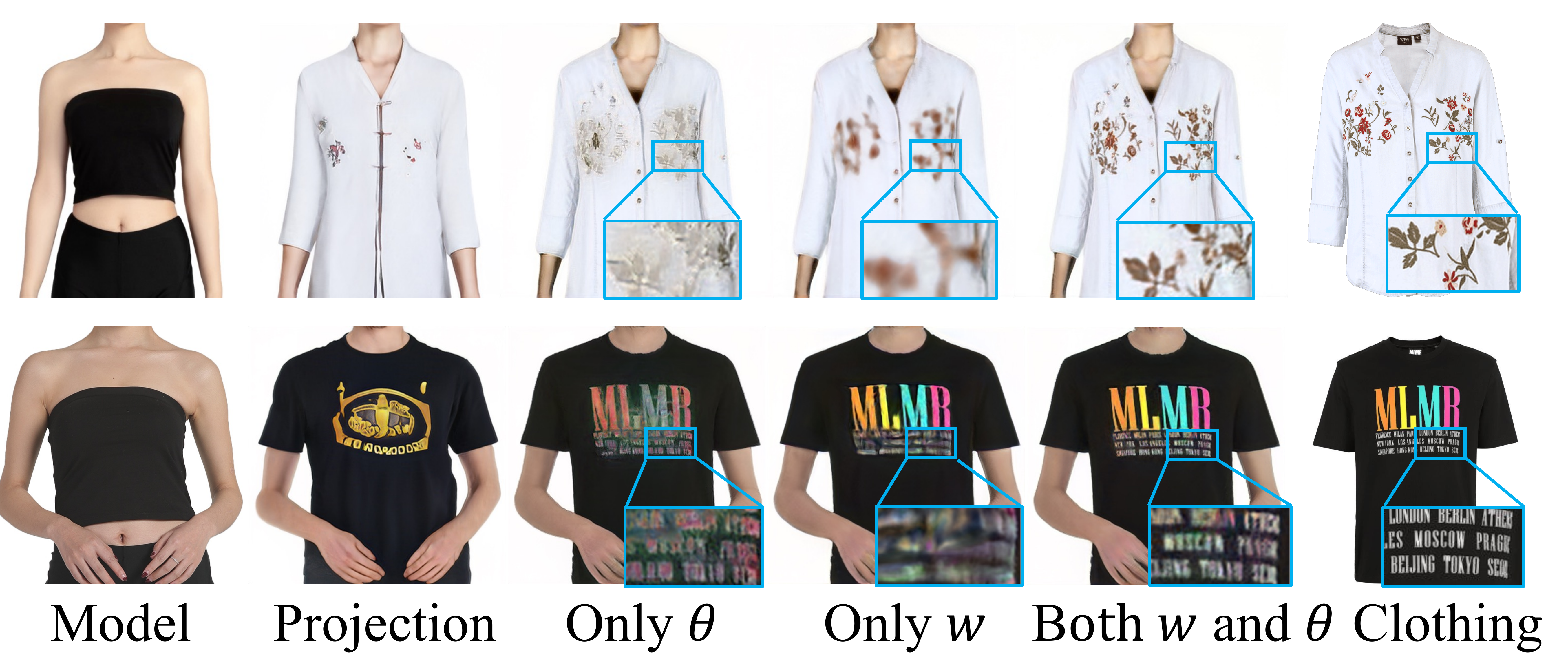}%\vspace{-0.2cm}
  \caption{Optimizing $\bm{\theta}$ alone recovers better texture details.
Optimizing $\w$ alone recovers better semantic information such as the overall color and shape of clothing characteristics. Optimizing $\bm{\theta}$ and $\w$ at the same time gets the optimal results.}\label{fig: figure_without_w_or_n}%\vspace{-0.2cm}
\end{figure}
\begin{figure}[t]
  \centering
  \includegraphics[width=1\linewidth]{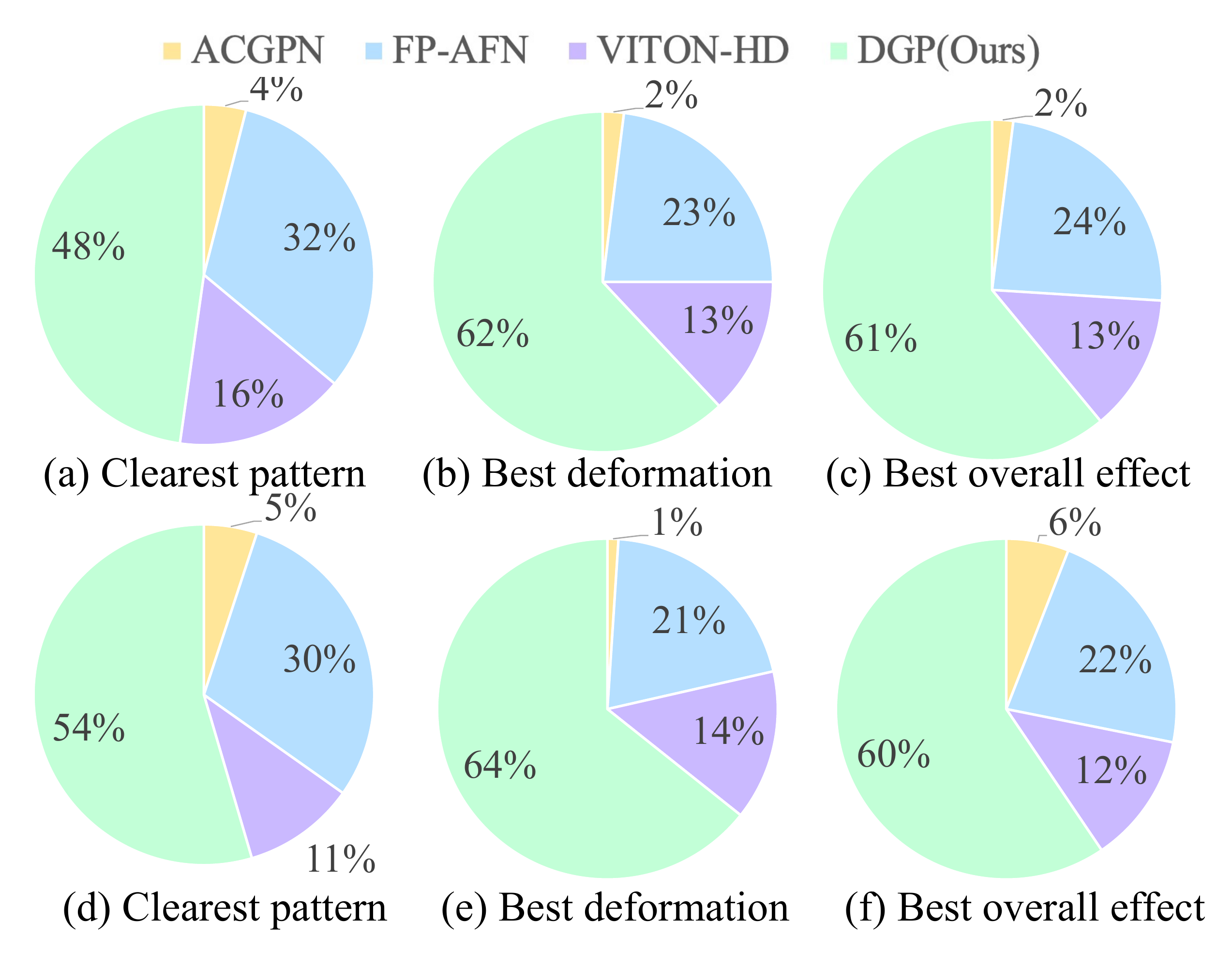}%\vspace{-0.25cm}
  \caption{User study on CMI ((a), (b), (c)) and MPV ((d), (e), (f)) datasets. The proposed weakly-supervised method outperforms all supervised competitors significantly over all three aspects.}\label{fig: figure_user_study_CMI}%\vspace{-0.4cm}
\end{figure}

\begin{figure*}[t]
  \centering
  \includegraphics[width=1\linewidth]{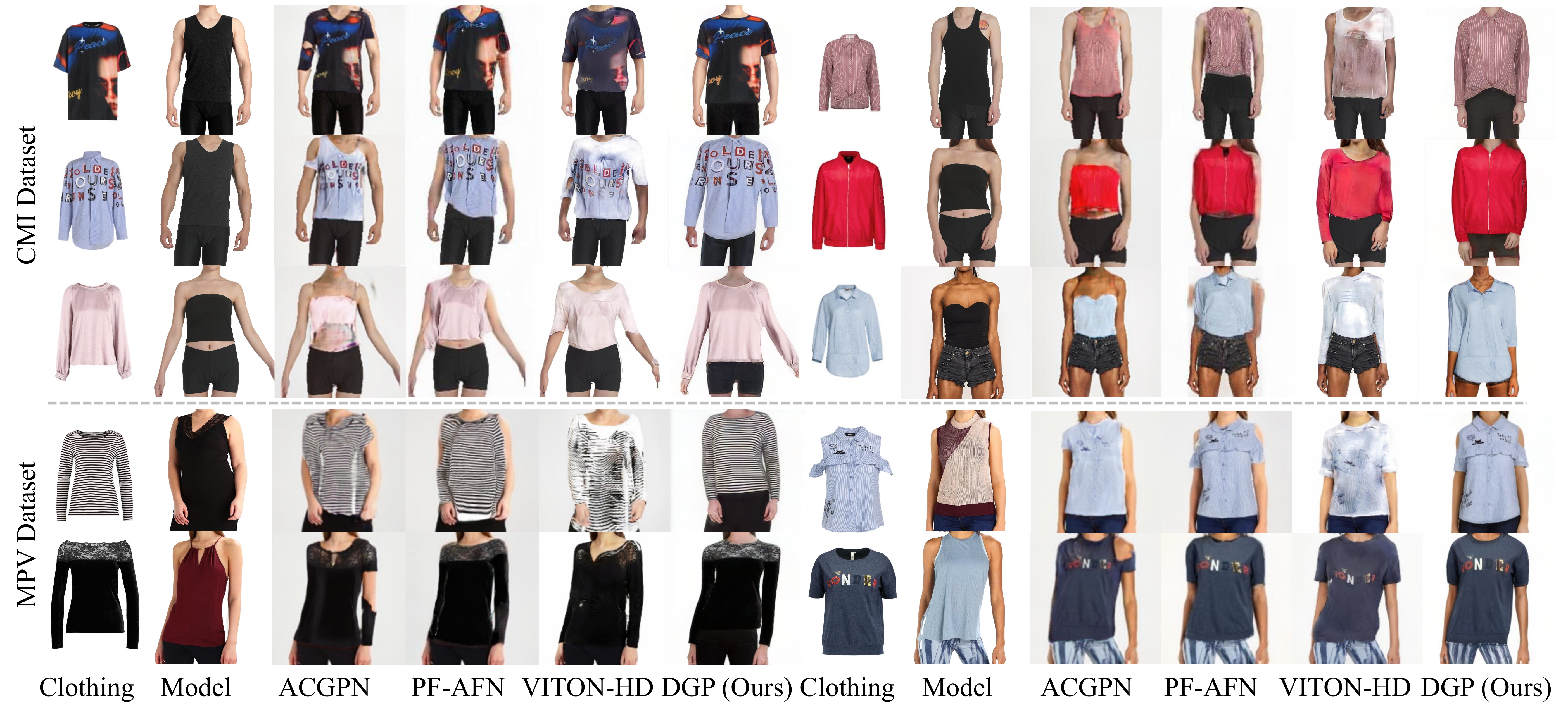}%\vspace{-0.25cm}
  \caption{Comparison on the CMI and MPV datasets. The supervised competitor methods are basically less appealing, and perform especially poorly on complicated clothing like coats.}\label{fig: compare_CMI}%\vspace{-0.25cm}
\end{figure*}

\section{Experiments}
In this section we evaluate the proposed \textit{weakly-supervised} DGP method from four different aspects, and compare it with several \textit{supervised} state-of-the-art competitors. Sec. \ref{sec:ablation} justifies the effect of the optimization components of the DGP method. Sec. \ref{sec:CMI} evaluates the performance on clothing model generation of the DGP method against some supervised competitors trained on paired image data. Here we focus on three state-of-the-art supervised methods, VITON-HD \cite{choi2021viton}, PF-AFN \cite{ge2021parser,han2019clothflow}, and ACGPN \cite{yang2020towards}. For experiments on all of those supervised methods, we use the pretrained models provided in their official repositories \cite{vitonhdcode,acpgncode,pfancode}. Sec. \ref{sec:MPV} further evaluates the performance on the MPV \cite{dong2019towards} dataset which is very similar to the training data \cite{han2018viton} of competitor methods. It is worthwhile to mention that their original training set VITON \cite{han2018viton} is no longer legal for academic use, thus MPV \cite{dong2019towards} may be the most pleasant dataset we can find for the competitor methods. Sec. \ref{sec:robust} evaluates the robustness of the proposed method against mistakes in the preprocessing. Throughout this paper, the optimizations of both semantic and pattern searches are terminated after 1,000 steps of projected gradient descent. Supplementary materials include table recording hyper-parameter selection of training and optimization objectives.

\subsection{Ablation Study}\label{sec:ablation}
In this section we conduct extensive experiments to verify the two optimization components of the DGP method. 
% \paragraph{Necessity of the projection}
% Fig. \aiur{fig:necessity of projection} reports the wearing results with and without projection module. The first one rows report the results of our proposed DGP methods, while the second rows report the results of removing the projection module. In this setting, we start the semantic search from the average point $\bm{\mu}$ of $\mathcal{W}+$ space, and conduct pattern search after it. We find that removing the projection module immediately ruins the results of semantic search, and yields poor overall results.  
%\paragraph{Necessity of the semantic and pattern search}
As shown in Fig. \ref{fig: figure_without_w_or_n}, optimizing $\theta$ alone recovers better texture details, but is less effective in reconstructing semantic information.
Optimizing $w$ alone, on the contrary, recovers better semantic information such as the overall color and shape, but poorer details of patterns and letters. Optimizing $\theta$ and $w$ at the same time yields the optimal results.

\begin{table}[t]
\caption{
    Numerical metrics of DGP, ACGPN, PF-AFN, and VITON-HD on CMI and MPV datasets. $\downarrow$ indicates lower is better.
  }%\vspace{-0.2cm}
  \label{tab:metrics}
  \centering
  \begin{tabular}{lcccc}
		\toprule
		\multirow{2}{*}{\textbf{Methods}} & \multicolumn{2}{c}{\textbf{CMI}} & \multicolumn{2}{c}{\textbf{MPV}} \\
		 \cmidrule(lr){2-3}\cmidrule(lr){4-5}& \FID & \SWD & \FID & \SWD\\
		\midrule
		%$512\times512$ resolution & & & &\\
		ACGPN&137.9 &121.3&81.1 &90.4\\
		PF-AFN &97.3&76.7&67.8&67.1\\
		VITON-HD &87.5&56.1&\textbf{40.6}&52.7\\
		DGP (Ours) &\textbf{51.6}&\textbf{22.4}&48.4&\textbf{36.7}\\
% 		\midrule
% 		$128\times128$ resolution & & & &\\
	
% 		\cmidrule(lr){1-1}ACGPN&115.7 &68.4&48.0 &39.2\\
% 		PF-AFN &86.6&29.4&49.5&\textbf{24.9}\\
% 		VITON-HD &95.0&27.4&\textbf{44.7}&25.6\\
% 		DGP (Ours) &\textbf{56.5}&\textbf{18.8}&46.8&29.9\\
		\bottomrule
	\end{tabular} %\vspace{-0.25cm}
\end{table}

\subsection{Clothing Model Generation}\label{sec:CMI}
The real scene clothing model generation task demands algorithms to handle unseen model and clothing images from unknown distributions. To evaluate algorithms under this scenario, we conduct experiments on the CMI benchmark dataset introduced in Sec. \ref{sec:task}, which is unavailable for all methods during training. For each model image of the CMI dataset, we randomly pick up a garment image from the 1,881 clothing images of CMI. It yields a testing set of 2,348 model and clothing pairs. All qualitative and quantitative evaluations are conducted on the 2,348 image pairs. The results are reported in Fig. \ref{fig: figure_user_study_CMI}, \ref{fig: compare_CMI}, and  Tab. \ref{tab:metrics}.

\paragraph{Qualitative Comparison} Fig. \ref{fig: compare_CMI} reports the qualitative comparison on the CMI dataset. The results reflect the advantages of the proposed method from three aspects. First, while not trained on the CMI dataset, the proposed method still works properly, with realistic synthesis. The competitors are overall less satisfactory, and work poorly in cases of complicated unseen clothes. Second, the proposed method synthesizes much clear patterns, while competitor methods often blur the patterns. Third, the proposed method can handle complicated clothing like coats, but competitor methods often fail in these cases.

% \begin{figure*}[h]
%   \centering
%   \includegraphics[width=1\linewidth]{compare_MPV.png}\vspace{-0.25cm}
%   \caption{Comparison on the MPV dataset. The proposed method yields far more realistic wearing results, together with preciser pattern reconstruction than all supervised competitors. It is worthwhile to mention that MPV is collected from the same source as the training data of all competitor methods, while the proposed method never sees those data.}\label{fig: compare_MPV}\vspace{-0.5cm}
% \end{figure*}

\paragraph{Quantitative Comparison} To quantitatively compare DGP with its competitors, we also measure the Fr\'echect Inception Distance (FID) \cite{heusel2017gans} and Sliced Wasserstein Distance (SWD) \cite{deshpande2018generative,kolouri2018sliced,kolouri2019generalized} of result images. While the CMI dataset does not contain ground-truth data, here the testing set of the previously mentioned E-Shop Fashion dataset is used as the reference images. All images are cropped to the same region and then resized to $512\times512$ resolution for a fair comparison. The results are reported in Tab. \ref{tab:metrics}. A user study is further conducted on the visual quality of the wearing results from three aspects: 1) which method generates the clearest pattern; 2) which method generates the most realistic wearing; 3) and which method generates the best overall effect. The results are reported in Fig. \ref{fig: figure_user_study_CMI}. Both numerical metrics and user study confirm the superiority of the proposed method. Details of user study are presented in supplementary materials.

\subsection{Comparison on MPV Dataset}\label{sec:MPV}
To challenge the proposed DGP method in an unfair setting, we further compare the results of all those methods on the MPV dataset. The MPV dataset is collected from the same source as the VITON dataset, which is the training set of the competitor methods and is no longer available due to legal issues. While DGP is not trained on MPV or VITON, the superiority in this scenario can be even more attractive. We pick 1,476 image pairs of person and clothing from the MPV dataset to construct the testing set. We report qualitative comparison in Fig. \ref{fig: compare_CMI}, numerical metrics of FID and SWD in Tab. \ref{tab:metrics}, and user study results in Fig. \ref{fig: figure_user_study_CMI}. Here an independent sampling of 1,476 images from MPV is used as the reference images to compute FID and SWD. The proposed method still maintains advantages in most aspects, and yields the same appealing results as in the CMI dataset.

\begin{figure}[h]
  \centering
  \includegraphics[width=1\linewidth]{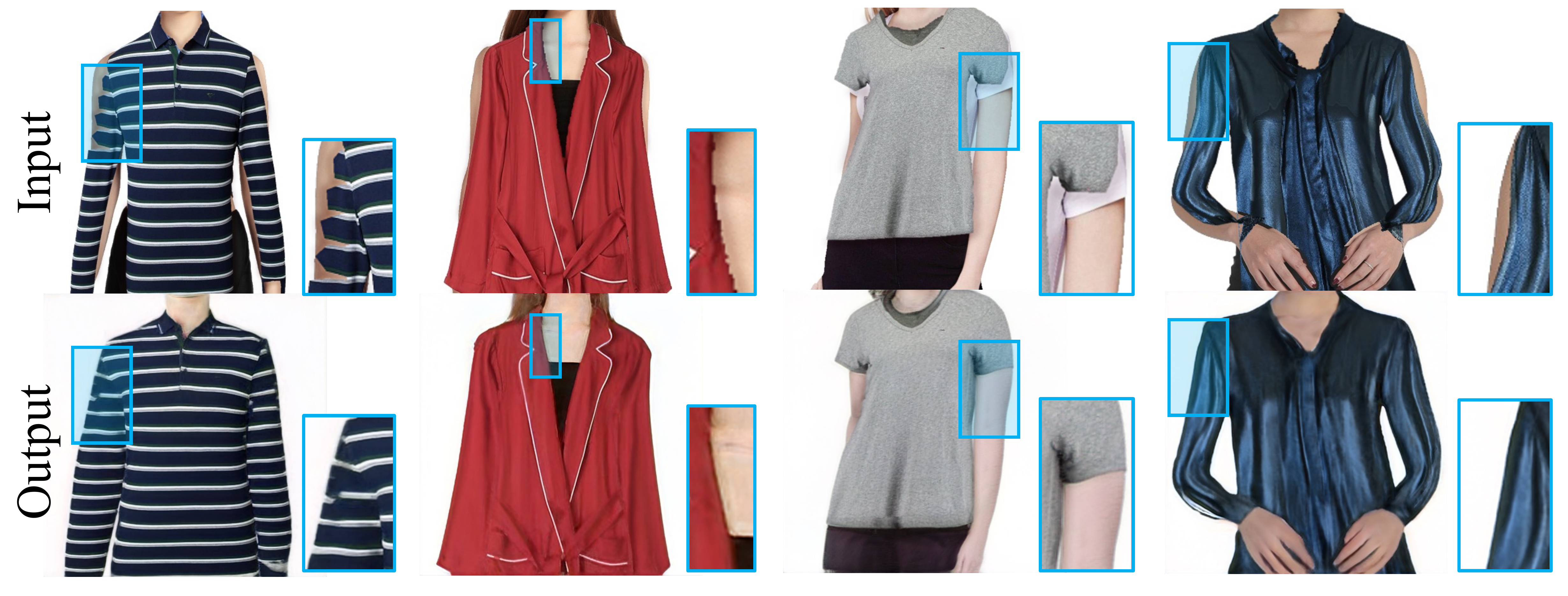}%\vspace{-0.3cm}
  \caption{Robustness of the DGP method. The mistakes like missing parts of clothing, wrong key point alignments, and zigzag clothing boundaries are easily corrected in the final results.}\label{fig: figure_robust}%\vspace{-0.5cm}
\end{figure}

\subsection{Robustness of DGP}\label{sec:robust}
 The imagination ability in Sec. \ref{sec:projection} of the projector is very appealing for the clothing model generation. So this section further investigates how this ability can help the DGP method overcome mistakes in the preprocessing period. We deliberately feed the DGP method with flawed rough alignment images, such as missing parts of clothing, wrong key point alignments, and zigzag clothing boundaries. We then observe how the DGP method will perform with those mistakes. The results are reported in Fig. \ref{fig: figure_robust}, which confirm that DGP can easily correct these tiny mistakes, and yield realistic synthesis in the final results.

\section{Limitations}
\begin{figure}[h]%\vspace{-0.2cm}
  \centering
  \includegraphics[width=1\linewidth]{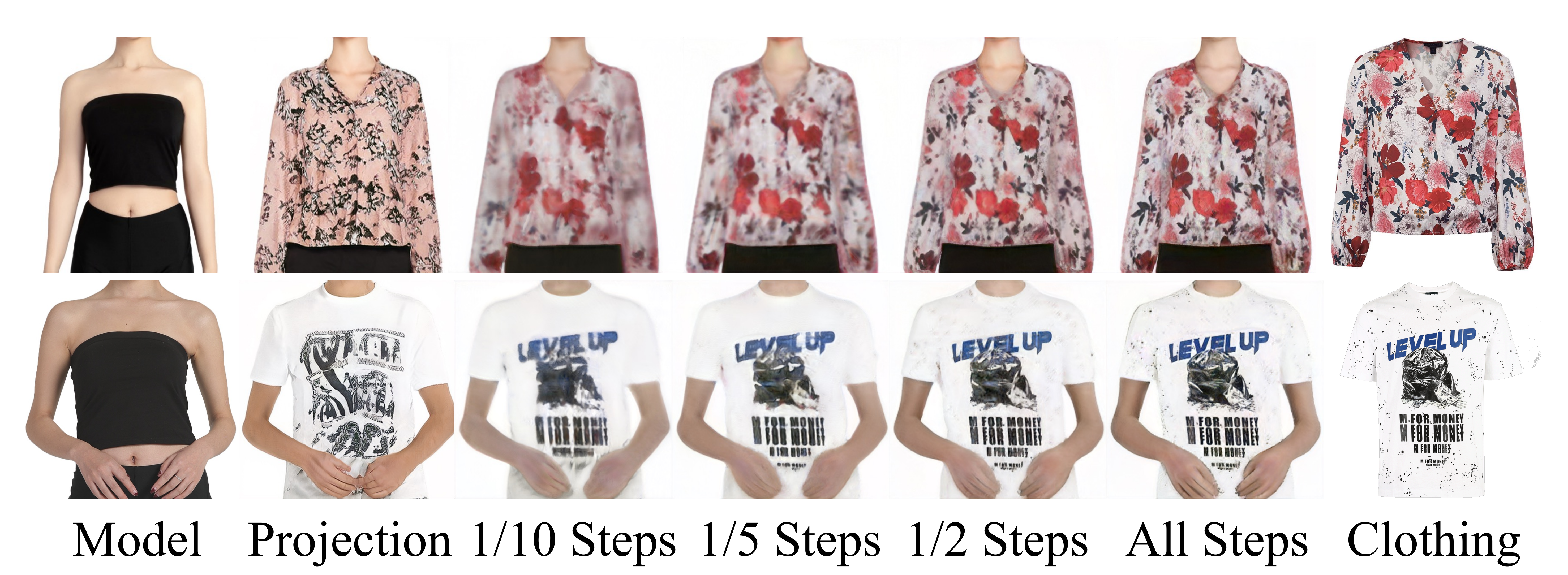}%\vspace{-0.35cm}
  \caption{Performance under different optimization steps. Best viewed in large size.}\label{fig: figure_inversion_time}%\vspace{-0.5cm}
\end{figure}

The major limitation of the DGP algorithm is the time cost of semantic and pattern searches. The current setting costs around 1 minute to finish the wearing of a single batch of data with one GTX 1080Ti GPU. Shortening the optimization steps may lower the overall performance, as reported in Fig. \ref{fig: figure_inversion_time}. It is acceptable for clothing model generation where we do not need real-time and interactive computation of results, but accelerating the computation can certainly benefit the method to apply on broader occasions. Also, the common difficulty of existing SOTA VTO methods in handling extremely complicated poses still remains in the proposed method. %\vspace{-0.2cm}

\section{Conclusion}
This paper studies the clothing model generation problem for online clothing retails. We propose a weakly supervised method to ease the demands of paired training data in typical virtual try-on algorithms. The proposed method casts the problem of warping clothing images to models' bodies into projecting a rough alignment of them onto the knowledge space of a pretrained StyleGAN. Extensive experiments demonstrate the superiority of our unpaired methods over several SOTA competitors trained with paired data. Future studies will focus on easing time consumption and increasing generality on extremely complicated poses of models.

{\small
\bibliographystyle{ieee_fullname}
\bibliography{egbib}

\begin{thebibliography}{10}\itemsep=-1pt

\bibitem{abdal2019image2stylegan}
Rameen Abdal, Yipeng Qin, and Peter Wonka.
\newblock Image2{StyleGAN}: How to embed images into the {StyleGAN} latent
  space?
\newblock In {\em Int. Conf. Comput. Vis.}, pages 4432--4441, 2019.

\bibitem{abdal2020image2stylegan++}
Rameen Abdal, Yipeng Qin, and Peter Wonka.
\newblock Image2{StyleGAN}++: How to edit the embedded images?
\newblock In {\em IEEE Conf. Comput. Vis. Pattern Recog.}, pages 8296--8305,
  2020.

\bibitem{albahar2021pose}
Badour AlBahar, Jingwan Lu, Jimei Yang, Zhixin Shu, Eli Shechtman, and Jia-Bin
  Huang.
\newblock Pose with style: Detail-preserving pose-guided image synthesis with
  conditional {StyleGAN}.
\newblock {\em arXiv preprint arXiv:2109.06166}, 2021.

\bibitem{alexa2000rigid}
Marc Alexa, Daniel Cohen-Or, and David Levin.
\newblock As-rigid-as-possible shape interpolation.
\newblock In {\em Proceedings of the 27th Annual Conference on Computer
  Graphics and Interactive Techniques}, pages 157--164, 2000.

\bibitem{boyd2004convex}
Stephen Boyd, Stephen~P Boyd, and Lieven Vandenberghe.
\newblock {\em Convex optimization}.
\newblock Cambridge university press, 2004.

\bibitem{brouet2012design}
Remi Brouet, Alla Sheffer, Laurence Boissieux, and Marie-Paule Cani.
\newblock Design preserving garment transfer.
\newblock {\em ACM Trans. Graph.}, 31(4):Article--No, 2012.

\bibitem{vitonhdcode}
Seunghwan Choi, Sunghyun Park, Minsoo Lee, and Jaegul Choo.
\newblock {VITON-HD} - official pytorch implementation.
\newblock \url{https://github.com/shadow2496/VITON-HD}, 2021.

\bibitem{choi2021viton}
Seunghwan Choi, Sunghyun Park, Minsoo Lee, and Jaegul Choo.
\newblock {VITON-HD}: High-resolution virtual try-on via misalignment-aware
  normalization.
\newblock In {\em IEEE Conf. Comput. Vis. Pattern Recog.}, pages 14131--14140,
  2021.

\bibitem{chung2001course}
Kai~Lai Chung and Kailai Zhong.
\newblock {\em A course in probability theory}.
\newblock Academic press, 2001.

\bibitem{deng2009imagenet}
Jia Deng, Wei Dong, Richard Socher, Li-Jia Li, Kai Li, and Li Fei-Fei.
\newblock Imagenet: A large-scale hierarchical image database.
\newblock In {\em IEEE Conf. Comput. Vis. Pattern Recog.}, pages 248--255.
  Ieee, 2009.

\bibitem{deshpande2018generative}
Ishan Deshpande, Ziyu Zhang, and Alexander~G Schwing.
\newblock Generative modeling using the sliced {W}asserstein distance.
\newblock In {\em IEEE Conf. Comput. Vis. Pattern Recog.}, pages 3483--3491,
  2018.

\bibitem{dong2019towards}
Haoye Dong, Xiaodan Liang, Xiaohui Shen, Bochao Wang, Hanjiang Lai, Jia Zhu,
  Zhiting Hu, and Jian Yin.
\newblock Towards multi-pose guided virtual try-on network.
\newblock In {\em Int. Conf. Comput. Vis.}, pages 9026--9035, 2019.

\bibitem{dong2019fw}
Haoye Dong, Xiaodan Liang, Xiaohui Shen, Bowen Wu, Bing-Cheng Chen, and Jian
  Yin.
\newblock F{W}-{GAN}: Flow-navigated warping {GAN} for video virtual try-on.
\newblock In {\em Int. Conf. Comput. Vis.}, pages 1161--1170, 2019.

\bibitem{dong2018boosting}
Yinpeng Dong, Fangzhou Liao, Tianyu Pang, Hang Su, Jun Zhu, Xiaolin Hu, and
  Jianguo Li.
\newblock Boosting adversarial attacks with momentum.
\newblock In {\em IEEE Conf. Comput. Vis. Pattern Recog.}, pages 9185--9193,
  2018.

\bibitem{gao2021shape}
Xin Gao, Zhenjiang Liu, Zunlei Feng, Chengji Shen, Kairi Ou, Haihong Tang, and
  Mingli Song.
\newblock Shape controllable virtual try-on for underwear models.
\newblock In {\em ACM Int. Conf. Multimedia}, pages 563--572, 2021.

\bibitem{ge2021parser}
Yuying Ge, Yibing Song, Ruimao Zhang, Chongjian Ge, Wei Liu, and Ping Luo.
\newblock Parser-free virtual try-on via distilling appearance flows.
\newblock In {\em IEEE Conf. Comput. Vis. Pattern Recog.}, pages 8485--8493,
  2021.

\bibitem{pfancode}
Yuying Ge, Yibing Song, Ruimao Zhang, Chongjian Ge, Wei Liu, and Ping Luo.
\newblock Parser-free virtual try-on via distilling appearance flows.
\newblock \url{https://github.com/geyuying/PF-AFN}, 2021.

\bibitem{ge2019deepfashion2}
Yuying Ge, Ruimao Zhang, Xiaogang Wang, Xiaoou Tang, and Ping Luo.
\newblock Deepfashion2: A versatile benchmark for detection, pose estimation,
  segmentation and re-identification of clothing images.
\newblock In {\em IEEE Conf. Comput. Vis. Pattern Recog.}, pages 5337--5345,
  2019.

\bibitem{goodfellow2014generative}
Ian Goodfellow, Jean Pouget-Abadie, Mehdi Mirza, Bing Xu, David Warde-Farley,
  Sherjil Ozair, Aaron Courville, and Yoshua Bengio.
\newblock Generative adversarial nets.
\newblock {\em Adv. Neural Inform. Process. Syst.}, 27, 2014.

\bibitem{guan2012drape}
Peng Guan, Loretta Reiss, David~A Hirshberg, Alexander Weiss, and Michael~J
  Black.
\newblock Drape: Dressing any person.
\newblock {\em ACM Trans. Graph.}, 31(4):1--10, 2012.

\bibitem{guo2019simple}
Chuan Guo, Jacob Gardner, Yurong You, Andrew~Gordon Wilson, and Kilian
  Weinberger.
\newblock Simple black-box adversarial attacks.
\newblock In {\em Proc. Int. Conf. Machine Learning}, pages 2484--2493. PMLR,
  2019.

\bibitem{han2019clothflow}
Xintong Han, Xiaojun Hu, Weilin Huang, and Matthew~R Scott.
\newblock Clothflow: A flow-based model for clothed person generation.
\newblock In {\em Int. Conf. Comput. Vis.}, pages 10471--10480, 2019.

\bibitem{han2018viton}
Xintong Han, Zuxuan Wu, Zhe Wu, Ruichi Yu, and Larry~S Davis.
\newblock {VITON}: An image-based virtual try-on network.
\newblock In {\em IEEE Conf. Comput. Vis. Pattern Recog.}, pages 7543--7552,
  2018.

\bibitem{hauswiesner2013virtual}
Stefan Hauswiesner, Matthias Straka, and Gerhard Reitmayr.
\newblock Virtual try-on through image-based rendering.
\newblock {\em IEEE Trans. Vis. Comput. Graph.}, 19(9):1552--1565, 2013.

\bibitem{he2016deep}
Kaiming He, Xiangyu Zhang, Shaoqing Ren, and Jian Sun.
\newblock Deep residual learning for image recognition.
\newblock In {\em IEEE Conf. Comput. Vis. Pattern Recog.}, pages 770--778,
  2016.

\bibitem{heusel2017gans}
Martin Heusel, Hubert Ramsauer, Thomas Unterthiner, Bernhard Nessler, and Sepp
  Hochreiter.
\newblock {GAN}s trained by a two time-scale update rule converge to a local
  {N}ash equilibrium.
\newblock {\em Adv. Neural Inform. Process. Syst.}

\bibitem{igarashi2005rigid}
Takeo Igarashi, Tomer Moscovich, and John~F Hughes.
\newblock As-rigid-as-possible shape manipulation.
\newblock {\em ACM Trans. Graph.}, 24(3):1134--1141, 2005.

\bibitem{issenhuth2020not}
Thibaut Issenhuth, J{\'e}r{\'e}mie Mary, and Cl{\'e}ment Calauzenes.
\newblock Do not mask what you do not need to mask: a parser-free virtual
  try-on.
\newblock In {\em Eur. Conf. Comput. Vis.}, pages 619--635. Springer, 2020.

\bibitem{jacobson2011bounded}
Alec Jacobson, Ilya Baran, Jovan Popovic, and Olga Sorkine.
\newblock Bounded biharmonic weights for real-time deformation.
\newblock {\em ACM Trans. Graph.}, 30(4):78, 2011.

\bibitem{jae2019viton}
Hyug Jae~Lee, Rokkyu Lee, Minseok Kang, Myounghoon Cho, and Gunhan Park.
\newblock {LA-VITON}: a network for looking-attractive virtual try-on.
\newblock In {\em Int. Conf. Comput. Vis.}, pages 0--0, 2019.

\bibitem{karras2018progressive}
Tero Karras, Timo Aila, Samuli Laine, and Jaakko Lehtinen.
\newblock Progressive growing of {GANs} for improved quality, stability, and
  variation.
\newblock In {\em Int. Conf. Learn. Represent.}, 2018.

\bibitem{karras2020training}
Tero Karras, Miika Aittala, Janne Hellsten, Samuli Laine, Jaakko Lehtinen, and
  Timo Aila.
\newblock Training generative adversarial networks with limited data.
\newblock {\em arXiv preprint arXiv:2006.06676}, 2020.

\bibitem{karras2019style}
Tero Karras, Samuli Laine, and Timo Aila.
\newblock A style-based generator architecture for generative adversarial
  networks.
\newblock In {\em IEEE Conf. Comput. Vis. Pattern Recog.}, pages 4401--4410,
  2019.

\bibitem{karras2020analyzing}
Tero Karras, Samuli Laine, Miika Aittala, Janne Hellsten, Jaakko Lehtinen, and
  Timo Aila.
\newblock Analyzing and improving the image quality of {StyleGAN}.
\newblock In {\em IEEE Conf. Comput. Vis. Pattern Recog.}, pages 8110--8119,
  2020.

\bibitem{kolouri2019generalized}
Soheil Kolouri, Kimia Nadjahi, Umut Simsekli, Roland Badeau, and Gustavo~K
  Rohde.
\newblock Generalized sliced {W}asserstein distances.
\newblock {\em arXiv preprint arXiv:1902.00434}, 2019.

\bibitem{kolouri2018sliced}
Soheil Kolouri, Gustavo~K Rohde, and Heiko Hoffmann.
\newblock Sliced {W}asserstein distance for learning {G}aussian mixture models.
\newblock In {\em IEEE Conf. Comput. Vis. Pattern Recog.}, pages 3427--3436,
  2018.

\bibitem{lancaster2005chi}
Henry~Oliver Lancaster and Eugene Seneta.
\newblock Chi-square distribution.
\newblock {\em Encyclopedia of Biostatistics}, 2, 2005.

\bibitem{laurent2000adaptive}
Beatrice Laurent and Pascal Massart.
\newblock Adaptive estimation of a quadratic functional by model selection.
\newblock {\em Annals of Statistics}, pages 1302--1338, 2000.

\bibitem{lewis2021tryongan}
Kathleen~M Lewis, Srivatsan Varadharajan, and Ira Kemelmacher-Shlizerman.
\newblock {TryOnGAN}: body-aware try-on via layered interpolation.
\newblock {\em ACM Trans. Graph.}, 40(4):1--10, 2021.

\bibitem{lewis2021vogue}
Kathleen~M Lewis, Srivatsan Varadharajan, and Ira Kemelmacher-Shlizerman.
\newblock {VOGUE}: Try-on by {StyleGAN} interpolation optimization.
\newblock {\em arXiv preprint arXiv:2101.02285}, 2021.

\bibitem{lin2014microsoft}
Tsung-Yi Lin, Michael Maire, Serge Belongie, James Hays, Pietro Perona, Deva
  Ramanan, Piotr Doll{\'a}r, and C~Lawrence Zitnick.
\newblock Microsoft coco: Common objects in context.
\newblock In {\em Eur. Conf. Comput. Vis.}, pages 740--755. Springer, 2014.

\bibitem{madry2017towards}
Aleksander Madry, Aleksandar Makelov, Ludwig Schmidt, Dimitris Tsipras, and
  Adrian Vladu.
\newblock Towards deep learning models resistant to adversarial attacks.
\newblock {\em arXiv preprint arXiv:1706.06083}, 2017.

\bibitem{mezirow1978perspective}
Jack Mezirow.
\newblock Perspective transformation.
\newblock {\em Adult education}, 28(2):100--110, 1978.

\bibitem{minar2020cp}
Matiur~Rahman Minar, Thai~Thanh Tuan, Heejune Ahn, Paul Rosin, and Yu-Kun Lai.
\newblock C{P}-{VTON}+: Clothing shape and texture preserving image-based
  virtual try-on.
\newblock In {\em IEEE Conf. Comput. Vis. Pattern Recog. Workshops}, 2020.

\bibitem{neuberger2020image}
Assaf Neuberger, Eran Borenstein, Bar Hilleli, Eduard Oks, and Sharon Alpert.
\newblock Image based virtual try-on network from unpaired data.
\newblock In {\em IEEE Conf. Comput. Vis. Pattern Recog.}, pages 5184--5193,
  2020.

\bibitem{patel2020tailornet}
Chaitanya Patel, Zhouyingcheng Liao, and Gerard Pons-Moll.
\newblock Tailornet: Predicting clothing in 3{D} as a function of human pose,
  shape and garment style.
\newblock In {\em IEEE Conf. Comput. Vis. Pattern Recog.}, pages 7365--7375,
  2020.

\bibitem{pons2017clothcap}
Gerard Pons-Moll, Sergi Pujades, Sonny Hu, and Michael~J Black.
\newblock Clothcap: Seamless 4{D} clothing capture and retargeting.
\newblock {\em ACM Trans. Graph.}, 36(4):1--15, 2017.

\bibitem{richardson2021encoding}
Elad Richardson, Yuval Alaluf, Or Patashnik, Yotam Nitzan, Yaniv Azar, Stav
  Shapiro, and Daniel Cohen-Or.
\newblock Encoding in style: a {StyleGAN} encoder for image-to-image
  translation.
\newblock In {\em IEEE Conf. Comput. Vis. Pattern Recog.}, pages 2287--2296,
  2021.

\bibitem{rohmer2010animation}
Damien Rohmer, Tiberiu Popa, Marie-Paule Cani, Stefanie Hahmann, and Alla
  Sheffer.
\newblock Animation wrinkling: augmenting coarse cloth simulations with
  realistic-looking wrinkles.
\newblock {\em ACM Trans. Graph.}, 29(6):1--8, 2010.

\bibitem{santesteban2019learning}
Igor Santesteban, Miguel~A Otaduy, and Dan Casas.
\newblock Learning-based animation of clothing for virtual try-on.
\newblock In {\em Computer Graphics Forum}, volume~38, pages 355--366. Wiley
  Online Library, 2019.

\bibitem{sarkar2021style}
Kripasindhu Sarkar, Vladislav Golyanik, Lingjie Liu, and Christian Theobalt.
\newblock Style and pose control for image synthesis of humans from a single
  monocular view.
\newblock {\em arXiv preprint arXiv:2102.11263}, 2021.

\bibitem{shen2020interpreting}
Yujun Shen, Jinjin Gu, Xiaoou Tang, and Bolei Zhou.
\newblock Interpreting the latent space of {GAN}s for semantic face editing.
\newblock In {\em IEEE Conf. Comput. Vis. Pattern Recog.}, pages 9243--9252,
  2020.

\bibitem{simonyan2014very}
Karen Simonyan and Andrew Zisserman.
\newblock Very deep convolutional networks for large-scale image recognition.
\newblock {\em arXiv preprint arXiv:1409.1556}, 2014.

\bibitem{sun2019deep}
Ke Sun, Bin Xiao, Dong Liu, and Jingdong Wang.
\newblock Deep high-resolution representation learning for human pose
  estimation.
\newblock In {\em IEEE Conf. Comput. Vis. Pattern Recog.}, pages 5693--5703,
  2019.

\bibitem{tewari2020stylerig}
Ayush Tewari, Mohamed Elgharib, Gaurav Bharaj, Florian Bernard, Hans-Peter
  Seidel, Patrick P{\'e}rez, Michael Zollhofer, and Christian Theobalt.
\newblock Stylerig: Rigging {StyleGAN} for 3{D} control over portrait images.
\newblock In {\em IEEE Conf. Comput. Vis. Pattern Recog.}, pages 6142--6151,
  2020.

\bibitem{tov2021designing}
Omer Tov, Yuval Alaluf, Yotam Nitzan, Or Patashnik, and Daniel Cohen-Or.
\newblock Designing an encoder for {StyleGAN} image manipulation.
\newblock {\em ACM Trans. Graph.}, 40(4):1--14, 2021.

\bibitem{wang2018toward}
Bochao Wang, Huabin Zheng, Xiaodan Liang, Yimin Chen, Liang Lin, and Meng Yang.
\newblock Toward characteristic-preserving image-based virtual try-on network.
\newblock In {\em Eur. Conf. Comput. Vis.}, pages 589--604, 2018.

\bibitem{wold1987principal}
Svante Wold, Kim Esbensen, and Paul Geladi.
\newblock Principal component analysis.
\newblock {\em Chemometrics and intelligent laboratory systems}, 2(1-3):37--52,
  1987.

\bibitem{wu2021stylespace}
Zongze Wu, Dani Lischinski, and Eli Shechtman.
\newblock Stylespace analysis: Disentangled controls for {StyleGAN} image
  generation.
\newblock In {\em IEEE Conf. Comput. Vis. Pattern Recog.}, pages 12863--12872,
  2021.

\bibitem{yang2020towards}
Han Yang, Ruimao Zhang, Xiaobao Guo, Wei Liu, Wangmeng Zuo, and Ping Luo.
\newblock Towards photo-realistic virtual try-on by adaptively
  generating-preserving image content.
\newblock In {\em IEEE Conf. Comput. Vis. Pattern Recog.}, pages 7850--7859,
  2020.

\bibitem{acpgncode}
Han Yang, Ruimao Zhang, Xiaobao Guo, Wei Liu, Wangmeng Zuo, and Ping Luo.
\newblock Towards photo-realistic virtual try-on by adaptively generating
  preserving image content.
\newblock \url{https://github.com/minar09/ACGPN}, 2020.

\bibitem{yu2019vtnfp}
Ruiyun Yu, Xiaoqi Wang, and Xiaohui Xie.
\newblock V{TNFP}: An image-based virtual try-on network with body and clothing
  feature preservation.
\newblock In {\em Int. Conf. Comput. Vis.}, pages 10511--10520, 2019.

\bibitem{zou2019fashionai}
Xingxing Zou, Xiangheng Kong, Waikeung Wong, Congde Wang, Yuguang Liu, and Yang
  Cao.
\newblock Fashion{AI}: A hierarchical dataset for fashion understanding.
\newblock In {\em IEEE Conf. Comput. Vis. Pattern Recog. Workshops}, pages
  0--0, 2019.

\end{thebibliography}
}

\newcommand{\beginsupplementary}{%
        \setcounter{equation}{0}
        \renewcommand{\theequation}{S\arabic{equation}}%
        \setcounter{table}{0}
        \renewcommand{\thetable}{S\arabic{table}}%
        \setcounter{figure}{0}
        \renewcommand{\thefigure}{S\arabic{figure}}%
     }
     
\renewcommand{\contentsname}{\centering Appendix}

% \twocolumn[{
% \renewcommand\twocolumn[1][]{#1}
% \maketitle

% \thispagestyle{empty}
% \hypersetup{linkcolor=black}
% \tableofcontents
% }]

\appendix
\beginsupplementary

\onecolumn

\section*{A Proof to Theorem 1}
\addcontentsline{toc}{section}{A Proof to Theorem 1}

\begin{theorem}
Assume that $\mathcal{W}+$ follows the multi-variable Gaussian distribution, then the output of the projector $\bm{P}$ will always fall in the high-density region of $\mathcal{W}+$, which is an $n$-dimensional ellipse $\mathcal{E}$ with axes $\q_1,...,\q_n$, and axis lengths $\psi \sigma_1^{\frac{1}{2}},...,\psi \sigma_n^{\frac{1}{2}}$. Rigorously, let $\omega_{n-1}$ denote the volume of the $n-1$ dimensional unit ball, for a random sample $\w$ from $\mathcal{W}+$, the possibility of it outside $\mathcal{E}$ is
\begin{equation}\label{eq:th1}
\begin{aligned}
    \mathbb{P}(\w \notin \mathcal{E})
    =\frac{1}{(2\pi)^{\frac{n}{2}}}\int_{\psi}^{\infty}\omega_{n-1}r^{n-1}e^{-\frac{1}{2}r^2}dr,
\end{aligned}
\end{equation}
which drops to zero drastically as $\psi$ grows larger; and for an arbitrary input $\bm{x}$, we have
\begin{equation}\label{eq:ellipse}
    \bm{P}(\bm{x})\in\mathcal{E}=\{\w:(\w-\bm{\mu})^T\bm{\Sigma}^{-1}(\w-\bm{\mu})\leq\psi^2\}.
\end{equation}
\end{theorem}
\newtheorem{remark}{Remark}

% \begin{remark}
% Note that, there are two typos in Eq. (2) \& (4) in the context of the paper, they should be
% \begin{equation}
%     \w_0=Tr(\bm{q_1}s_1\sqrt{\sigma_1}+...\bm{q}_n s_n\sqrt{\sigma_n})+\bm{\mu}=\bm{Q}\bm{\Lambda}^{\frac{1}{2}}Tr(\bm{s})+\bm{\mu},
% \end{equation}
% \begin{equation}
%     \w=\bm{P}(\bm{x})=\bm{Q}\bm{\Lambda}^{\frac{1}{2}}Tr(\bm{s})+\bm{\mu},
% \end{equation}
% correspondingly.
% \end{remark}

Proof to this theorem yields three parts: proving that the output of the projector $\bm{P}$ always falls in $\mathcal{E}$ (Eq. (\ref{eq:ellipse})), proving Eq. (\ref{eq:th1}), and demonstrating that it decreases to zero as $\psi$ grows.

\subsection*{A.1 Proof to Eq. (\ref{eq:ellipse})}
\addcontentsline{toc}{subsection}{A.1 Proof to Eq. (\ref{eq:ellipse})}
Recall the computation of the Projection in Sec. 4.1. Given a rough alignment image $\x_a$, we first train an encoder $\bm{E}$ to get the strength code $\bm{s}$
\begin{equation}
    \bm{s}=\bm{E}(\x_a).
\end{equation}
Then we truncate the strength code $\bm{s}$ to an ellipse centered at $\bm{\mu}$ with radius $\psi$:
\begin{equation}
    \w_0=Tr(\bm{q_1}s_1\sqrt{\sigma_1}+...\bm{q}_n s_n\sqrt{\sigma_n})+\bm{\mu}=\bm{Q}\bm{\Lambda}^{\frac{1}{2}}Tr(\bm{s})+\bm{\mu},
\end{equation}
where $Tr$ is a truncation operator with cutoff coefficient $\psi>0$ such that
\begin{align}
Tr(\bm{v})=
\left\{\begin{array}{ll}
     \bm{v}, \Vert \bm{v}\Vert_2<\psi,\\
     \frac{\bm{v}}{\Vert\bm{v}\Vert_2}\psi, \Vert \bm{v}\Vert_2\geq\psi.
\end{array}\right.
\end{align}

For any $\bm{s}$, it is then easy to see that
\begin{equation}\label{eq:ellipse2}
    (\bm{Q}^{-1}\bm{\Lambda}^{-\frac{1}{2}}(\w_0-\bm{\mu}))^T(\bm{Q}^{-1}\bm{\Lambda}^{-\frac{1}{2}}(\w_0-\bm{\mu}))=(\w-\bm{\mu})^T\bm{Q}\bm{\Lambda}^{-1}\bm{Q}^T(\w-\bm{\mu})=Tr(\bm{s})^T Tr(\bm{s})\leq \psi^2.
\end{equation}
As $\bm{Q}$ is computed from PCA decomposition, we have
\begin{equation}\label{eq:pca1}
    \bm{Q}^T\bm{Q}=\bm{I}\Leftrightarrow \bm{Q}^{-1}=\bm{Q}^T,
\end{equation}
\begin{equation}\label{eq:pca2}
    \bm{\Sigma}=\bm{Q}\bm{\Lambda}\bm{Q}^T.
\end{equation}
Take Eq. (\ref{eq:pca1}) \& (\ref{eq:pca2}) to (\ref{eq:ellipse2}), we then have
\begin{equation}
    (\w_0-\bm{\mu})^T\bm{\Sigma}^{-1}(\w_0-\bm{\mu})\leq\psi^2,
\end{equation}
which verifies Eq. (\ref{eq:ellipse}).

\subsection*{A.2 Proof to Eq. (\ref{eq:th1})}
\addcontentsline{toc}{subsection}{A.2 Proof to Eq. (\ref{eq:th1})}
Let assume that $\w\sim \mathcal{N}(\bm{\mu}_{\w},\bm{\Sigma}_{\w})$. The Law of Large Numbers \cite{chung2001course} then tells that, for a collection of \textit{i.i.d} sampling $\{\w_i\}_{i=1}^N$ from the $\mathcal{W}+$ space, we have
\begin{equation}
    \bm{\mu}=\frac{1}{N}\sum_{i=1}^N\w_i \rightarrow \bm{\mu}_{\w},\,\,as\,\,N\rightarrow\infty,
\end{equation}
\begin{equation}
    \bm{\Sigma}=\frac{1}{N-1}(\w_1-\bm{\mu},...,\w_n-\bm{\mu})^T(\w_1-\bm{\mu},...,\w_n-\bm{\mu})\rightarrow \bm{\Sigma}_{\w},\,\,as\,\,N\rightarrow\infty.
\end{equation}
Considering that we sample five million points to compute the $\bm{\mu}$ and $\bm{\Sigma}$, $N$ is large enough to let us assume that 
\begin{equation}
    \bm{\mu}=\bm{\mu}_{\w},\bm{\Sigma}=\bm{\Sigma}_{\w}.
\end{equation}
Thus we have $\w\sim\mathcal{N}(\bm{\mu},\bm{\Sigma})$. It is then easy to see that
\begin{equation}
    \hat{\w}=\sqrt{\bm{\Sigma}^{-1}}^T(\w-\bm{\mu})\sim \mathcal{N}(\mathcal{O},\mathcal{I}),
\end{equation}
where $\sqrt{\bm{\Sigma}^{-1}}\sqrt{\bm{\Sigma}^{-1}}^T=\bm{\Sigma}^{-1}$, $\bm{\Sigma}^{-1}$ is the inverse matrix of $\bm{\Sigma}$, and $\mathcal{N}(\mathcal{O},\mathcal{I})$ is the Standard Gaussian distribution.
Note that
\begin{equation}
     \w\notin\mathcal{E}\Leftrightarrow(\w-\bm{\mu})^T\bm{\Sigma}^{-1}(\w-\bm{\mu})>\psi^2\Leftrightarrow \hat{\w}^T\hat{\w}>\psi^2\Leftrightarrow \hat{\w}\notin B(0,\psi).
\end{equation}
Thus we have
\begin{gather}
    \mathbb{P}(\w\notin\mathcal{E})=\mathbb{P}(\hat{\w}\notin B(0,\psi)).
\end{gather}
As $\hat{\w}$ follows the Standard Gaussian, we know that the sum of squares of all its element follows the $n$-dimensional chi-square distribution \cite{chung2001course,lancaster2005chi},
\begin{equation}
    \hat{\w}^T\hat{\w}\sim \chi_n^2.
\end{equation}
Thus we have 
\begin{equation}
    \mathbb{P}(\w\notin\mathcal{E})=\mathbb{P}(\hat{\w}\notin B(0,\psi))=\mathbb{P}(\chi_n^2>\psi^2)=\frac{1}{(2\pi)^{\frac{n}{2}}}\int_{\psi}^{\infty}\omega_{n-1}r^{n-1}e^{-\frac{1}{2}r^2}dr.
\end{equation}

\subsection*{A.3 Proof to Eq. (\ref{eq:th1}) Decreasing to Zero}
This is easy to see as it is well known that the tail of chi-square distribution falls to zero quickly. Here we quote the tail bound of chi-square distribution in \cite{laurent2000adaptive} (Lemma 1 p1325), that 
\begin{equation}
    \mathbb{P}(\chi_n^2>n+2\sqrt{nt}+2t)<e^{-t}, \forall t\in\mathbb{R}_+.
\end{equation}
When $\psi$ is large enough, we further have
\begin{equation}
    \mathbb{P}(\chi_n^2>\psi^2)<e^{-\frac{\psi^2}{10}}.
\end{equation}
Both the above two estimations verify that the tail of chi-square distribution falls to zero drastically.

% \section*{B Ablation Study}
% \addcontentsline{toc}{section}{B Ablation Study}
% \subsection*{B.1 Project vs. No Projector}
% \addcontentsline{toc}{subsection}{B.1 Project vs. No Projector}
% To justify the effect of the proposed projector, we conduct ablation study to investigate the synthesis quality with and without the projector. Here we remove the projector and carry out semantic search and pattern search directly from the mean value $\bm{\mu}$ of the $\mathcal{W}$+ space. This strategy is used in some related works like Image2StyleGAN \cite{abdal2019image2stylegan,abdal2020image2stylegan++} and VOGUE \cite{lewis2021vogue}. The results are reported in Fig. \ref{fig: SM_no_projector_mask_n}. Removing the projector will immediately ruin the synthesis, yielding low quality results. Thus the projector is necessary for the proposed method to succeed.

% \begin{figure}[h]
%   \centering
%   \includegraphics[width=1\linewidth]{with_and_without_projector.png}
%   \caption{Results with and without using the projector.}\label{fig: SM_no_projector_mask_n}
% \end{figure}

\section*{B Numerical Measurements of Each DGP Components}
\addcontentsline{toc}{section}{B Numerical Measurements of Each DGP Components}
\begin{table}[h]
\caption{
    Numerical metrics of the Projection, Semantic Search, and Pattern Search steps of the DGP method on the CMI dataset, at the resolution of $512\times512$. $\downarrow$ indicates lower is better.
  }
  \label{tab:metrics_s}
  \centering
  \begin{tabular}{lcccc}
		\toprule
		Metrics & Projection & Semantic Search & Pattern Search & DGP (final output) \\
		\midrule
		\FID & 54.2 & 52.9 & 58.9 & 58.9 \\
		\SWD & 56.7 & 45.7 & 46.5 & 46.5 \\
		\bottomrule
	\end{tabular} 
\end{table}

\section*{C Rough Alignment}
\addcontentsline{toc}{section}{C Rough Alignment}

\begin{figure}[h]
  \centering
  \includegraphics[width=1\linewidth]{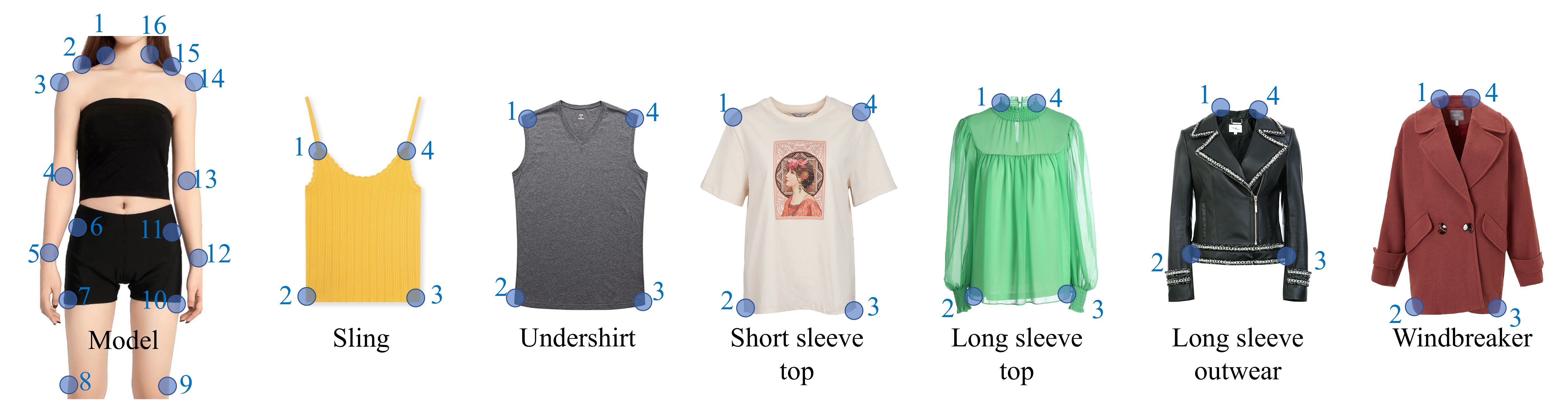}
  \caption{Definition of model key points and clothing key points with six basic categories.}\label{fig: SM_keypoints_model_cloth}
\end{figure}

The rough alignment step uses a collection of model key points together with a collection of clothes key points to compute the alignment. The collection of model key points include points of neck, shoulder, hip, elbow, and wrist. Among them, key points of elbow and wrist are inherited from the corresponding annotations in the COCO \cite{lin2014microsoft} dataset for human poses; key points of neck, shoulder, and hip are acquired from an API interface from an anonymous online AI platform. Fig. \ref{fig: SM_keypoints_model_cloth} and Tab. \ref{tab:model_keypoints} demonstrate more details of those key points. 
We feed images from the COCO dataset to the anonymous API interface to obtain the labels for neck, shoulder, and hip, and merge them with the original labels in COCO to obtain the ground truth key point annotations for training. 
An HRNet \cite{sun2019deep} model is trained on those images to predict these key points. The HRNet model reaches 67.2 AP on the test set after convergence. 

% 1 left neck
% 2 left collarbone
% 3 left shoulder
% 4 left elbow
% 5 left wrist
% 6 left hip
% 7 left thigh
% 8 left knee
% 9 right knee
% 10 right thigh
% 11 right hip
% 12 right wrist
% 13 right elbow
% 14 right shoulder
% 15 right collarbone
% 16 right neck

% \begin{table}[h]
%     \caption{Definition of model key points.}
%     \label{tab:model_keypoints}
%     \centering
%     \begin{tabular}{|l|l|l|l|}
%         \hline
%         Index & Name & Index & Name \\ \hline
%         1 & left neck & 9 & right knee \\ \hline
%         2 & left collarbone & 10 & right thigh \\ \hline
%         3 & left shoulder & 11 & right hip \\ \hline
%         4 & left elbow & 12 & right wrist \\ \hline
%         5 & left wrist & 13 & right elbow \\ \hline
%         6 & left hip & 14 & right shoulder \\ \hline
%         7 & left thigh & 15 & right collarbone \\ \hline
%         8 & left knee & 16 & right neck \\ \hline
%     \end{tabular}
% \end{table}

\begin{table}[h]
    \caption{Definition of model key points.}
    \label{tab:model_keypoints}
    \centering
    \begin{tabular}{llllllll}
        \toprule
        Index & Name & Index & Name & Index & Name & Index & Name \\ 
        \midrule
        1 & left neck & 5 & left wrist & 9 & right knee & 13 & right elbow \\ 
        2 & left collarbone & 6 & left hip & 10 & right thigh & 14 & right shoulder\\ 
        3 & left shoulder & 7 & left thigh &  11 & right hip & 15 & right collarbone \\ 
        4 & left elbow & 8 & left knee & 12 & right wrist & 16 & right neck \\ 
        \bottomrule
    \end{tabular}
\end{table}

\begin{table}[h]
    \caption{Definition of clothing key points.}
    \label{tab:clothing_keypoints}
    \centering
    \begin{tabular}{lllllllll}
        \toprule
        Category & Index & Name & Index & Name & Index & Name & Index & Name \\ 
        \midrule
        Sling & 1 & left collarbone & 2 & left hip & 3 & right hip & 4 & right collarbone\\ 
        Undershirt & 1 & left collarbone & 2 & left hip & 3 & right hip & 4 & right collarbone\\ 
        Short sleeve top & 1 & left shoulder & 2 & left hip & 3 & right hip & 4 & right shoulder\\ 
        Long sleeve top & 1 & left neck & 2 & left hip & 3 & right hip & 4 & right neck\\ 
        Long sleeve outwear & 1 & left neck & 2 & left hip & 3 & right hip & 4 & right neck\\ 
        Windbreaker & 1 & left neck & 2 & left hip & 3 & right hip & 4 & right neck\\ 
        \bottomrule
    \end{tabular}
\end{table}

The collection of clothes key points are inherited from the DeepFashion2 \cite{ge2019deepfashion2} dataset. Again, an HRNet model is trained on the DeepFashion2 dataset, with 56 AP on the test set after convergence. Key points such as neckline, shoulder, and bottom corner of the clothing are selected to perform the perspective transformation, the else are used to guide the As Rigid As Possible (ARAP) \cite{alexa2000rigid,igarashi2005rigid} transformation. Fig. \ref{fig: SM_keypoints_model_cloth} and Tab. \ref{tab:clothing_keypoints} illustrate this process.

\begin{figure}[h]
  \centering
  \includegraphics[width=1\linewidth]{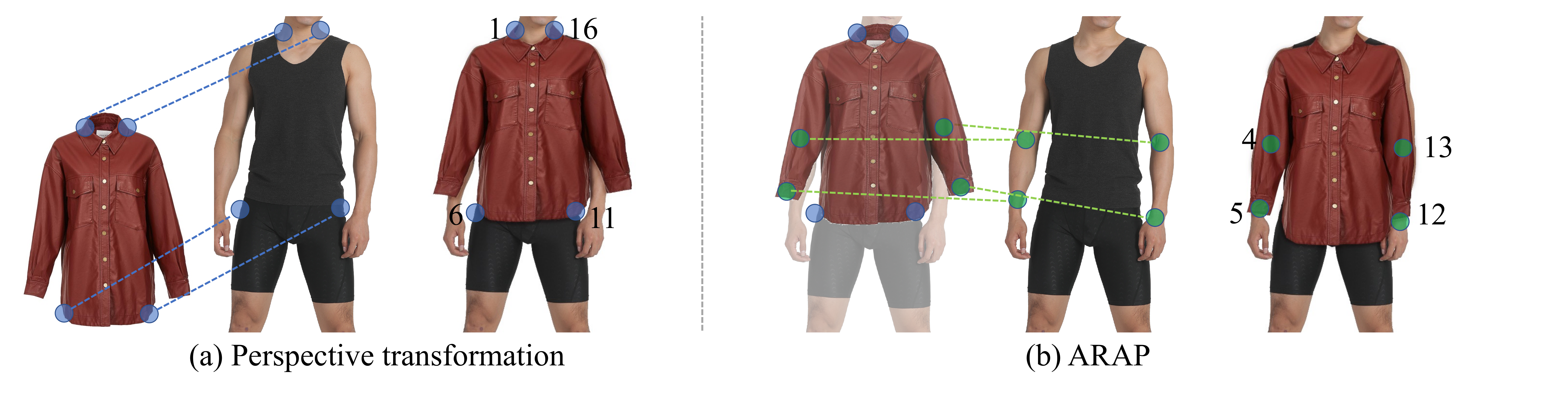}
  \caption{Illustration of (a) the perspective transformation and (b) the ARAP, the index numbers of the corresponding key points on the model have been marked.}\label{fig: SM_rough_alignment}
\end{figure}

After getting those key points, the rough alignment can be split into two phases: 1) a perspective transformation, followed by 2) an ARAP transformation. We use the OpenCV function \textit{cv2.getPerspectiveTransform} to compute the transformation matrix $\bm{A}$ that aligns the clothing key points at the left and right boundaries of the neckline and the bottom corner to the model key points at the left and right boundaries of the neck and the hip, correspondingly. Another OpenCV function \textit{cv2.WarpPerspective} is used to carry out the alignment of clothing and model image under transformation $\bm{A}$, and generate the result of the whole perspective transformation phase.
% to the key points at the left and right boundaries of the neckline and the bottom corner, correspondingly. 
Fig. \ref{fig: SM_rough_alignment} (a) illustrates this process. For different types of clothes, the key points we used admit subtle differences, Tab. \ref{tab:mapping_rule} reports those differences. Specifically, we omit the ARAP transformation for sleeveless and short-sleeve tops, thus the result of the perspective transformation is the final result of the rough alignment. For the long-sleeve type, we further use ARAP to align limbs, as shown in Fig. \ref{fig: SM_rough_alignment} (b). ARAP offers an interface to compute an energy minimized isotropy deformation given the offsets of couples of control points. Here we use the four skeleton points of the elbows and wrists of the clothing as the control points, and the four skeleton points of the elbows and wrists of the model image mentioned above as their destinations, correspondingly. Key points of the neck and hip, which have been aligned in the previous perspective transformation phase, are also added to the pool of control points as the immovable control points, which will be fixed during the ARAP deformation. Then the ARAP will roughly align the sleeves of the clothing to the arms of the model, and keep the body part of the clothing fixed. We take this as the final output of the rough alignment for long-sleeve categories such as jackets and sweaters.

\begin{table}[h]
    \caption{Mapping rules of the perspective transformation .}
    \label{tab:mapping_rule}
    \centering
    \begin{tabular}{llll}
        \toprule
        Clothing type & Points mapping rules (clothing-model) \\ 
         \midrule
        Sling & 1-2, 2-6, 3-11, 4-15  \\ 
        Undershirt & 1-2, 2-6, 3-11, 4-15  \\ 
        Short sleeve top & 1-3, 2-6, 3-11, 4-14  \\ 
        Long sleeve top & 1-1, 2-6, 3-11, 4-16  \\ 
        Long sleeve outwear & 1-1, 2-6, 3-11, 4-16  \\
        Windbreaker & 1-1, 2-6, 3-11, 4-16  \\ 
        \bottomrule
    \end{tabular}
\end{table}

\section*{D Attribute Classifier}
\addcontentsline{toc}{section}{D Attribute Classifier}
The optimization process of inversion on human face images often uses identity loss to help in recovering detailed information of the face. Similarly, the training loss for the Encoder includes the attribute similarity loss $\mathcal{L}_{attr}$, which is captured by a pretrained clothing attribute classifier $\bm{R}$ trained on the FashionAI dataset \cite{zou2019fashionai}. We select 7 category dimensions (neck, collar, lapel, neckline, sleeves length, skirt length, top length), 49 tags in total, from the FashionAI dataset, and train an attribute classification model based on the ResNet50 architecture. The model is trained on 4 Tesla V100 GPUS, with an mAP score of 0.95 and an accuracy score of 0.84 after convergence. 
The final convolution layer is taken as the feature space to compute the similarity between the generated image and the rough alignment image.

\begin{figure}[h]
  \centering
  \includegraphics[width=1\linewidth]{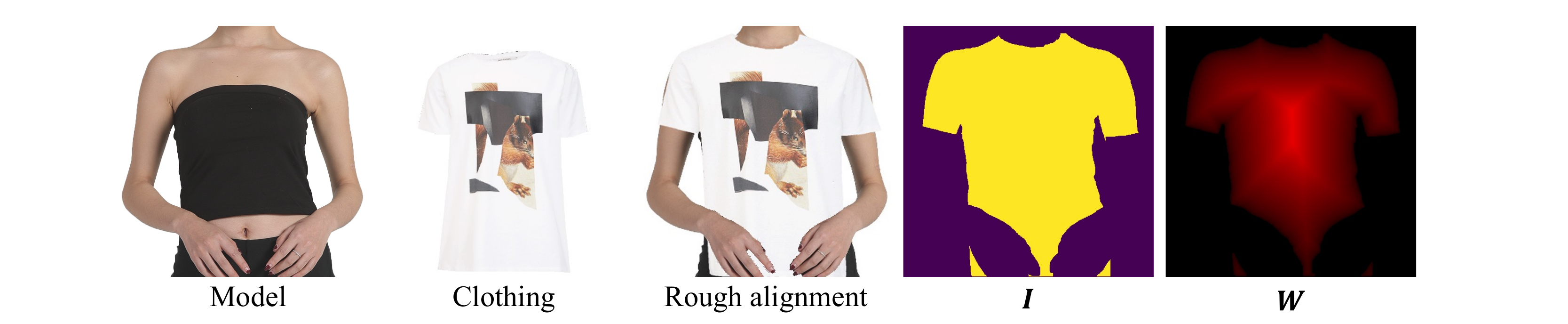}
  \caption{Dynamic spatial weight $\bm{W}$ in Semantic Search.}\label{fig: dynamic_W}
\end{figure}

\section*{E Dynamic Spatial Weight}
\addcontentsline{toc}{section}{E The E-Shop Fashion (ESF) Dataset}
Fig. \ref{fig: dynamic_W} illustrates the isosurface of the Dynamic Spatial Weight $\bm{W}$ in Semantic Search and Pattern Search, which is computed according to Eq. (18) of the paper context. Deeper red color denotes the higher weight, and the black background denotes zero weight. 
% $\bm{W}$ is generated according to the formula in the paper, and the shade of red indicates the weight size of the place. 
The distance function $d((i,j),\partial I)$ for an arbitrary pixel point in position $(i,j)$ is computed based on the erosion operation (\textit{cv2.erode}) in OpenCV Library. We recurrently conduct erosion operation to the region $I$ with $3\times3$ kernel size. In the $k$-th step, if the spatial position $(i,j)$ gets eroded, we set $d((i,j),\partial I)=k$. For spatial position $(i,j)$ at the boundary of the region $I$ or outside the region $I$, we set $d((i,j),\partial I)=0$.

\section*{F The E-Shop Fashion (ESF) Dataset }
\begin{figure}[h]
  \centering
  \includegraphics[width=1\linewidth]{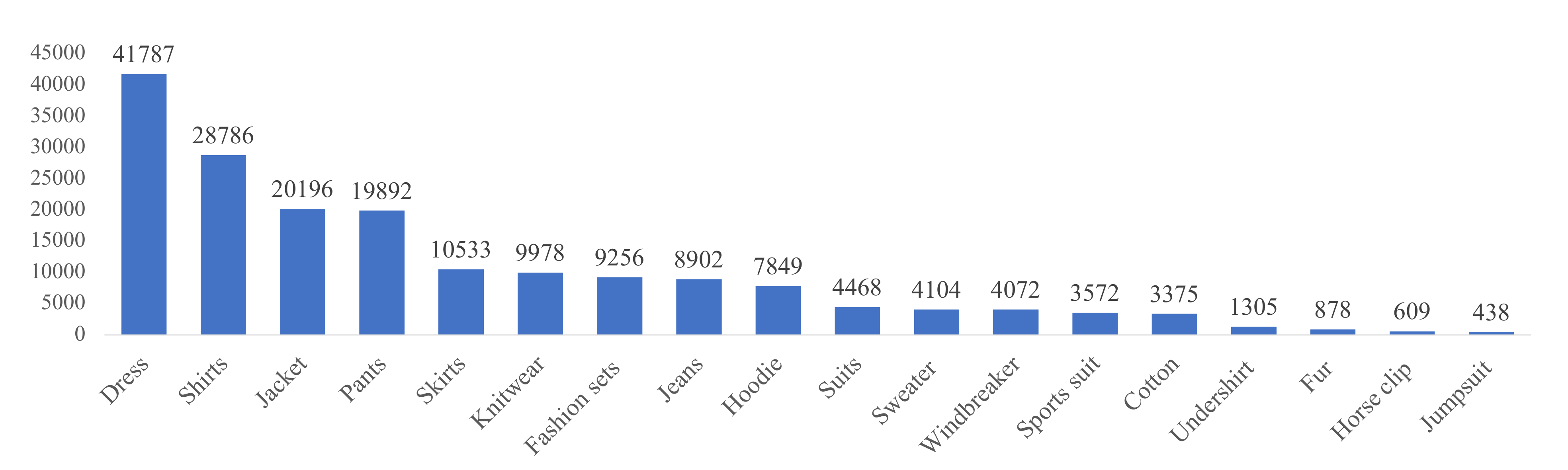}
  \caption{Number of each category on the ESF dataset. The total number is 180,000.}\label{fig: SM_sta_cloth}
\end{figure}

\begin{figure}[h]
  \centering
  \includegraphics[width=1\linewidth]{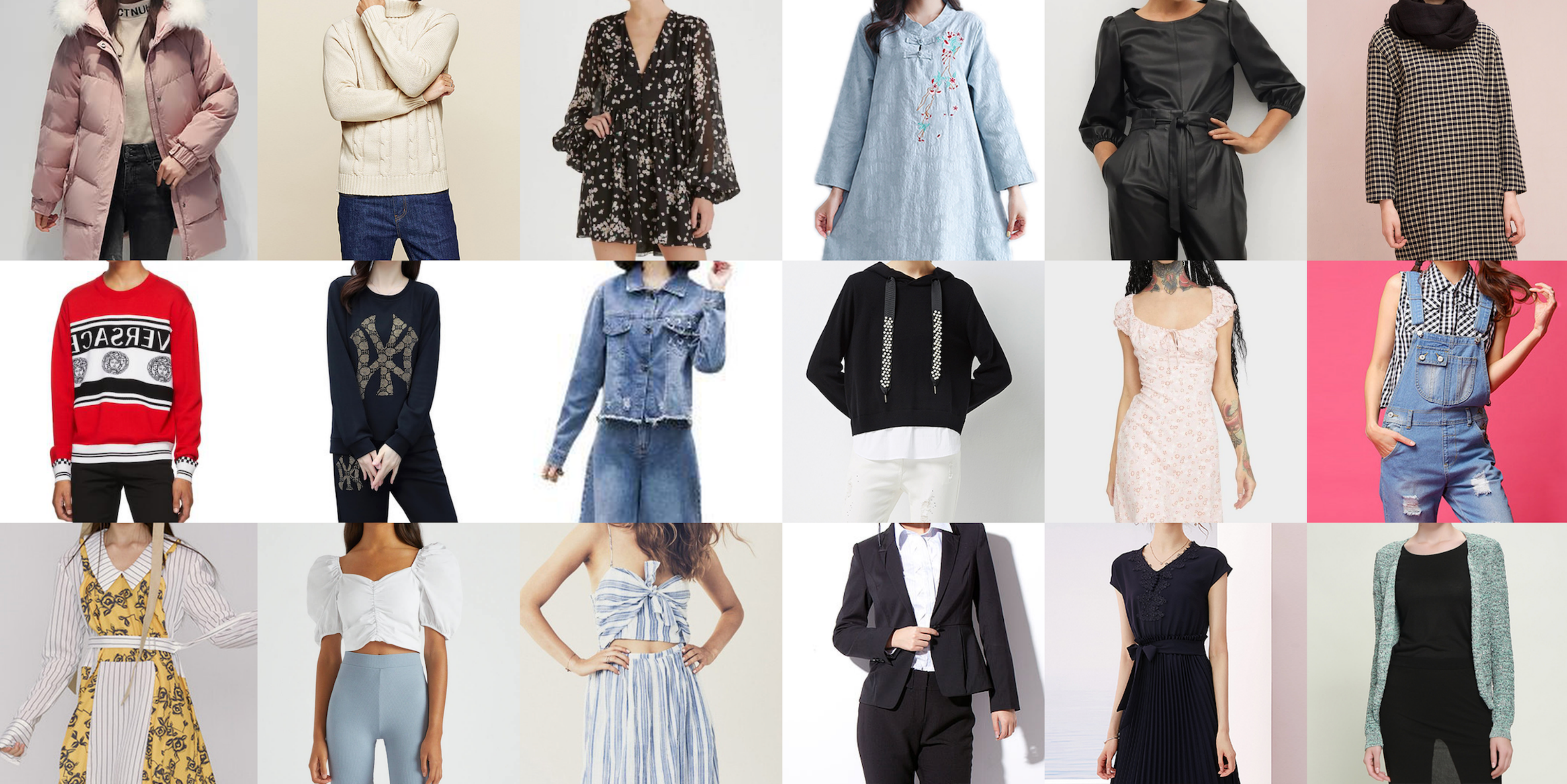}
  \caption{Examples of the ESF dataset.}\label{fig: SM_ESF}
\end{figure}

\addcontentsline{toc}{section}{F Dynamic Spatial Weight}
The E-Shop Fashion (ESF) dataset contains 180,000 clothing model images collected from an anonymous e-commercial website under legal circumstances. To ensure the purity of the dataset, we filter out images with complex backgrounds and models holding bags or other items. Each image contains only one full-body, frontal standing pose model.
A detection model is used to get the bounding box of the model, and are center aligned with resize and padding operations. The images are all cropped to the region between jaw and thigh, and resized to the resolution of 512 $\times$ 512. Fig. \ref{fig: SM_sta_cloth} gives the category names and the corresponding quantities, and Fig. \ref{fig: SM_ESF} shows some examples of the ESF dataset. The whole dataset contains 180,000 images, which is further split into 170,000 training samples and 10,000 testing samples. This dataset will be open-sourced later.

\begin{figure}[h]
  \centering
  \includegraphics[width=1\linewidth]{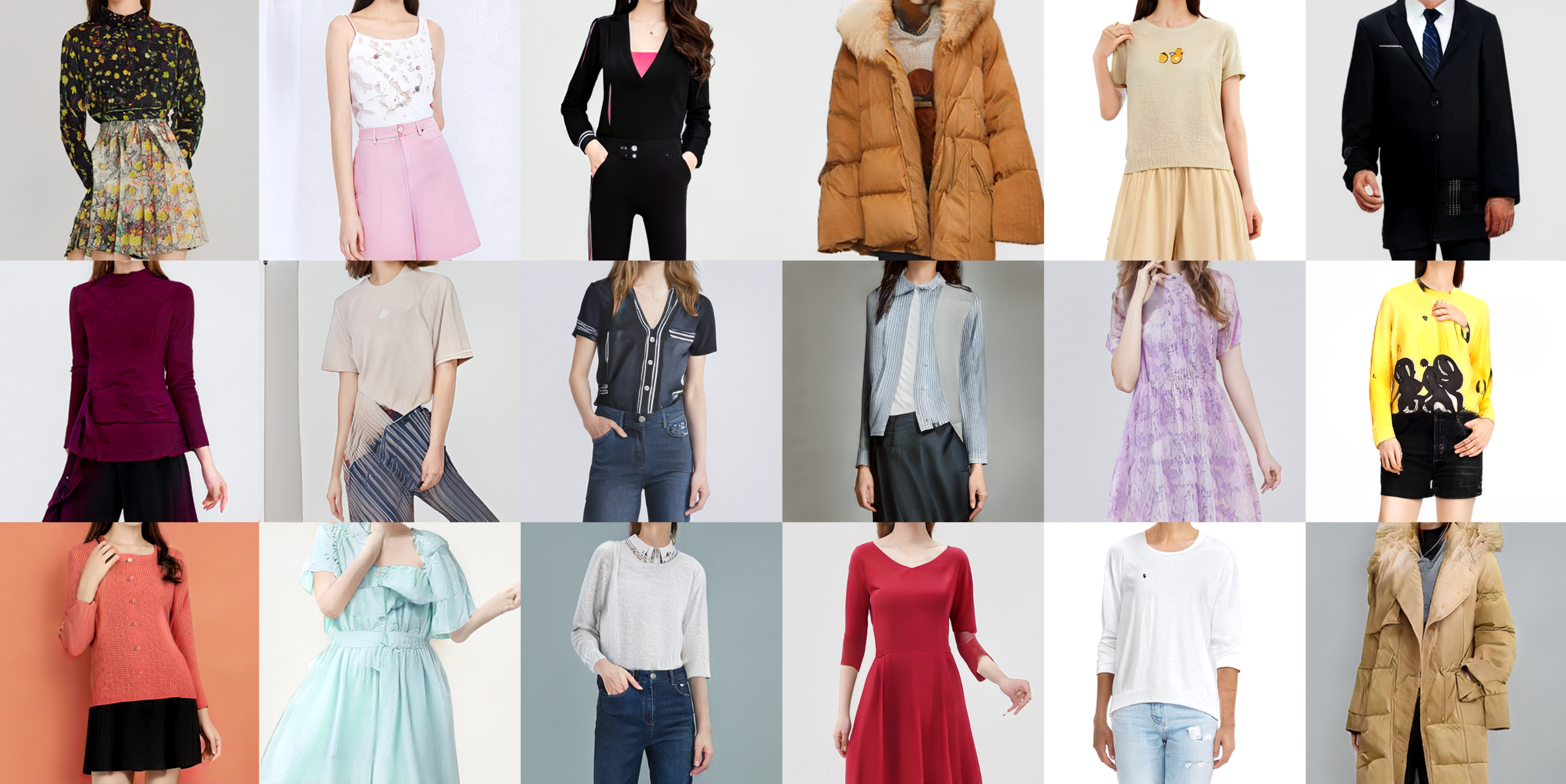}
  \caption{Fake images generated by StyleGAN2. The final FID is 2.16.}\label{fig: SM_fake_img}
\end{figure}

\section*{G The Commercial Model Image (CMI) Dataset}
\begin{figure}[h]
  \centering
  \includegraphics[width=1\linewidth]{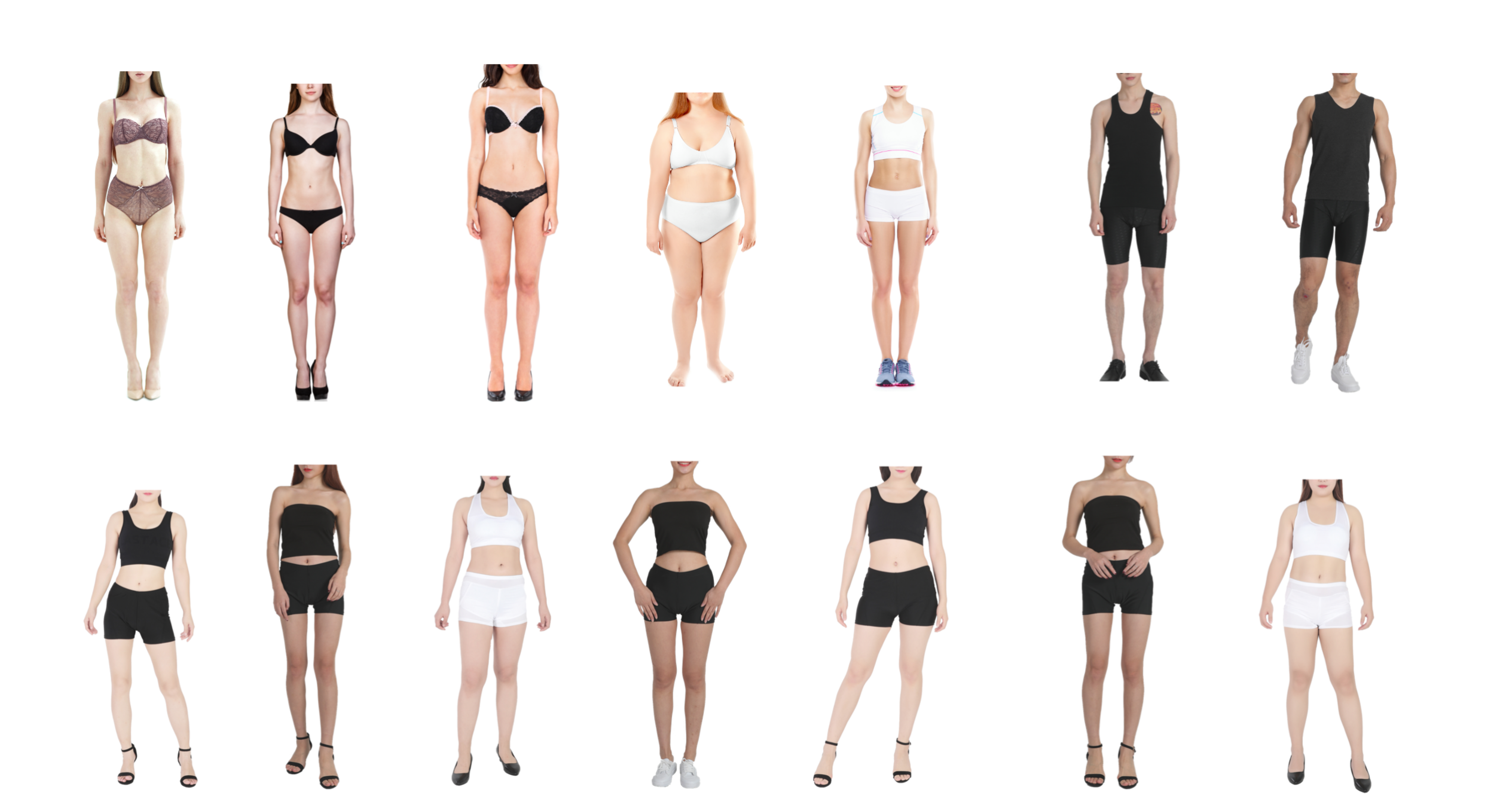}
  \caption{Examples of model images on the CMI dataset.}\label{fig: SM_CMI_dataset}
\end{figure}

\begin{figure}[h]
  \centering
  \includegraphics[width=1\linewidth]{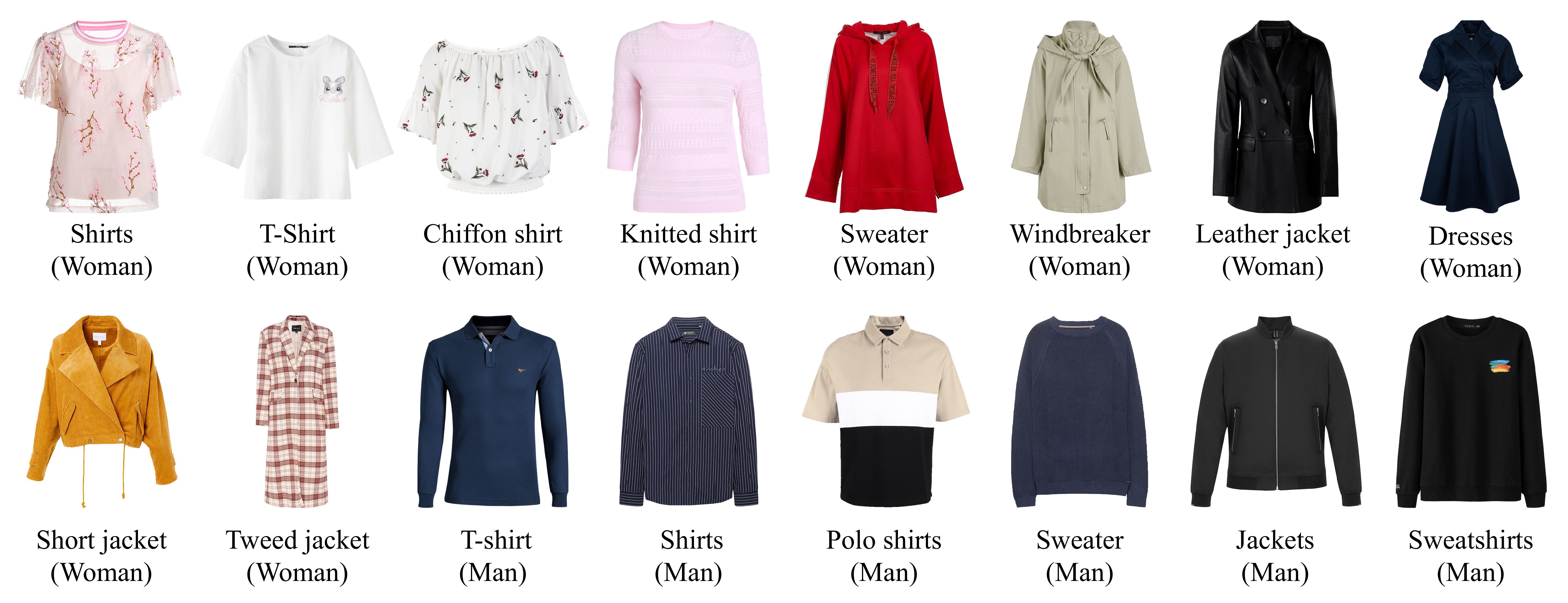}
  \caption{Examples of 16 categories of clothing images on the CMI dataset.}\label{fig: SM_CMI_dataset_cloth}
\end{figure}

\addcontentsline{toc}{section}{G The Commercial Model Image (CMI) Dataset}
The Commercial Model Image (CMI) dataset contains 2,348 images taken by 200 models on
underwear, including different genders, ages, body shapes,
and poses. All model images are taken in professional studios, and all models signed the confidentiality and authorization statement file. Fig. \ref{fig: SM_CMI_dataset} shows some models of the CMI dataset.
Another part of the CMI dataset is 1,881 clothing images with clean backgrounds from an anonymous e-commercial website under legal circumstances, which evenly contains 16 categories of top clothing, including 10 categories of women's tops and 6 categories of men's tops. Each category contains 120 images, excepting woman's leather jacket that only contains 81 images. 
Fig. \ref{fig: SM_CMI_dataset_cloth} shows examples of these 16 categories.

\section*{H License of the CMI Dataset}
\addcontentsline{toc}{section}{H License of the CMI Dataset}
Since the CMI dataset contains 2,348 images taken by 200 models on underwear, authorization and privacy issues must be considered. Each model is asked to sign a confidentiality and authorization statement file.
The contents of the file include authorization, confidentiality and privacy statement. This file named CONFIDENTIALITY AND AUTHORIZATION STATEMENT.pdf can be found in the zip archive, please read this file for more details.
In addition, to protect the models' privacy, the face areas are blurred for all the images. We cropped the face part during the pre-processing step to ensure that all figures in the paper and the supplementary materials do not contain faces.

% Women's shirts
% Women's T-shirt
% Lace shirt/chiffon shirt
% Women's knitted sweater
% Women's sweater / fleece shirt
% Women's windbreaker
% Women's leather jacket
% Women's dresses
% Women's short jacket
% Women's tweed jacket

% Men's T-shirt
% Men's shirts
% Men's polo shirts
% Men's knitted sweater/sweater
% Men's jackets
% Men's Sweatshirts

\section*{I User Study}
\addcontentsline{toc}{section}{I User Study}
We conduct user studies on our CMI benchmark and MPV \cite{dong2019towards} dataset by recruiting 50 volunteers.
The proposed DGP method is compared with three state-of-the-art supervised
methods, VITON-HD  \cite{choi2021viton}, PF-AFN \cite{ge2021parser,han2019clothflow} and ACGPN \cite{yang2020towards}. 
For each model image of the CMI dataset, we randomly pick up a garment image from the 1,881 clothing images of CMI. It yields a testing set of 2,348
model and clothing pairs. 
For the MPV dataset, We pick 1,476 image pairs of person and clothing to construct the testing set.
Each sample of the testing set contains six images, i.e., a target clothing, a reference person image, together with the try-on results of three compared methods and our DGP method. The order of the four try-on results is randomly shuffled. All 50 volunteers are required to mark all samples of the test set, and are asked to answer the following questions: 1) which method generates the clearest pattern; 2) which method generates the most realistic wearing; 3) and which method generates the
best overall effect. For each sample, the method with the highest number of votes will be the winner.

% \section{Broader Impact}

\section*{J Hyper-parameter Table}
\addcontentsline{toc}{section}{J Hyper-parameter Table}
Please refer to Tab. \ref{tab:hyper} for hyper-parameter selection of training and optimization objectives.

\begin{table}[!h]
\caption{
    Hyper-parameter selection of training and optimization objectives.
  }
  \label{tab:hyper}
  \centering
  \begin{tabular}{lcccc}
		\toprule
		Components & Loss & Learning Rate & Terminating Condition \\
		\midrule
        StyleGAN $\bm{G}$ &  The same as \cite{karras2020training}  & $2.5e-3$ &1,250,000 iterations\\
		StyleGAN $\bm{D}$ &  The same as \cite{karras2020training}  & $2.5e-3$ &1,250,000 iterations\\
		Projector $\bm{P}$ & $\lambda_p=1.0,\lambda_f=5e-5,\lambda_{attr}=5e-5,\lambda_{adv}=0.1,\psi=6.$ & $2e-5$ & 562,500 iterations\\
		Semantic Search & $\eta_p=1.0,\eta_f=5e-5,\eta_{attr}=5e-5,\eta_{adv}=1.0.$ & $1e-2$ & 1,000 PGD iterations\\
		Pattern Search & $\eta_p=1.0.$ & $1e-2$ & 1,000 PGD iterations\\
		\bottomrule
	\end{tabular}
\end{table}

\section*{K Numerical Results on 128 and 512 Resolutions}
\addcontentsline{toc}{section}{K Numerical Results on 128 and 512 Resolutions}
Please refer to Tab. \ref{tab:different_resolution} for numerical metrics of DGP, ACGPN, PF-AFN, and VITON-HD on CMI and MPV datasets. 

\begin{table}[!h]
\caption{
    Numerical metrics of DGP, ACGPN, PF-AFN, and VITON-HD on CMI and MPV datasets. $\downarrow$ indicates lower is better.
  }
  \label{tab:different_resolution}
  \centering
  \begin{tabular}{lcccc}
		\toprule
		\multirow{2}{*}{\textbf{Methods}} & \multicolumn{2}{c}{\textbf{CMI}} & \multicolumn{2}{c}{\textbf{MPV}} \\
		 \cmidrule(lr){2-3}\cmidrule(lr){4-5}& \FID & \SWD & \FID & \SWD\\
		\midrule
		$512\times512$ resolution & & & &\\
		\cmidrule(lr){1-1}ACGPN&137.9 &121.3&81.1 &90.4\\
		PF-AFN &97.3&76.7&67.8&67.1\\
		VITON-HD &87.5&56.1&\textbf{40.6}&52.7\\
		DGP (Ours) &\textbf{51.6}&\textbf{22.4}&48.4&\textbf{36.7}\\
		\midrule
		$128\times128$ resolution & & & &\\
	
		\cmidrule(lr){1-1}ACGPN&115.7 &68.4&48.0 &39.2\\
		PF-AFN &86.6&29.4&49.5&\textbf{24.9}\\
		VITON-HD &95.0&27.4&\textbf{44.7}&25.6\\
		DGP (Ours) &\textbf{56.5}&\textbf{18.8}&46.8&29.9\\
		\bottomrule
	\end{tabular} 
\end{table}

\section*{L Training Details of StyleGAN2}

\begin{figure}[t]
  \centering
  \includegraphics[width=0.5\linewidth]{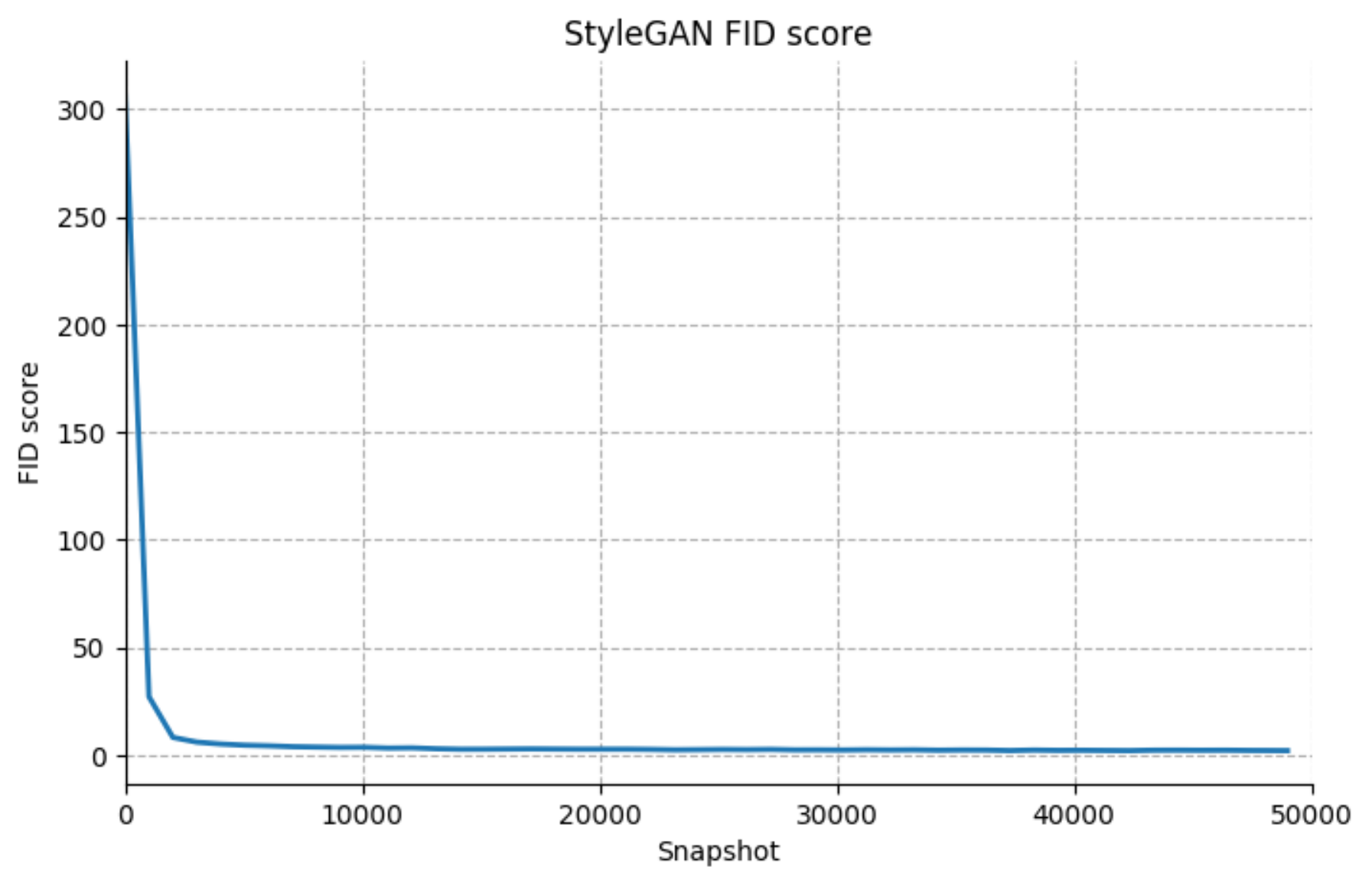}
  \caption{FID score during training.}\label{fig: stylegan_fid}
\end{figure}

\addcontentsline{toc}{section}{L Training Details of StyleGAN2}
The official StyleGAN2-ADA \cite{karras2020training} is used to train the StyleGAN2 model. We train StyleGAN2-ADA on 8 NVIDIA A100 Tensor Core GPUs, where we set the hyper-parameter \textit{batch\_size} as 64, \textit{ema\_kimg} as 20.0, and \textit{ema\_rampup} as 0.05. The resolution of the generated images is 512 $\times$ 512, the ADA options of brightness, contrast, and saturation are turned on, and horizontal flipping is allowed. The optimizer of the generator is Adam, the learning rate is 0.0025, and the discriminator settings are the same. The loss function remains the same as in the official StyleGAN2-ADA. The training stops with 48,988,000 images, with an early stop strategy, and the final FID is 2.16. The FID score during training is shown in Fig. \ref{fig: stylegan_fid}, and samples of generated images are shown in Fig. \ref{fig: SM_fake_img}.

\clearpage

\begin{figure}[h]
  \centering
  \includegraphics[width=0.9\linewidth]{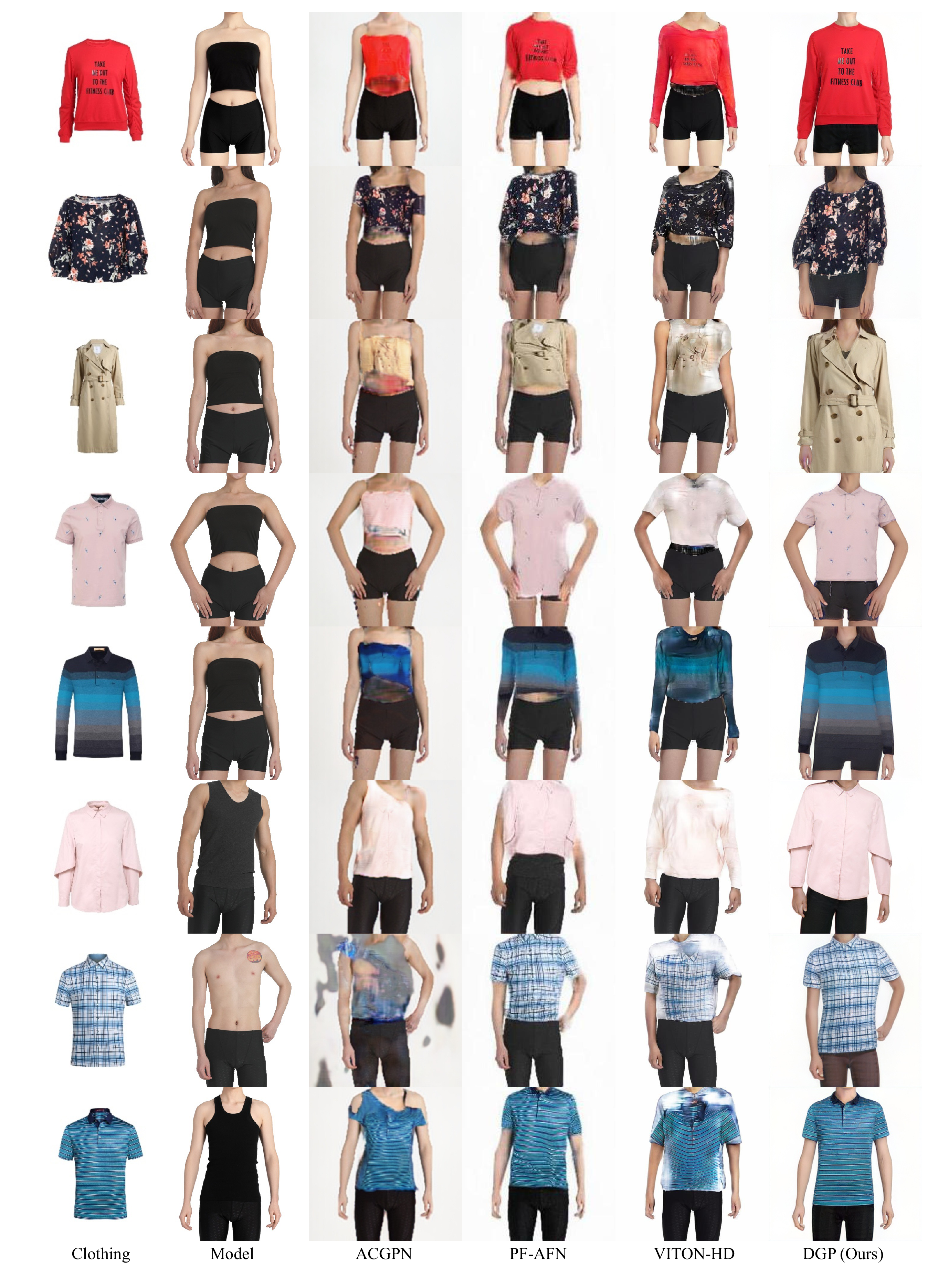}
  \caption{More visual results of qualitative comparison on the CMI dataset.}\label{fig: SM_compare_CMI}
\end{figure}

\begin{figure}[h]
  \centering
  \includegraphics[width=0.9\linewidth]{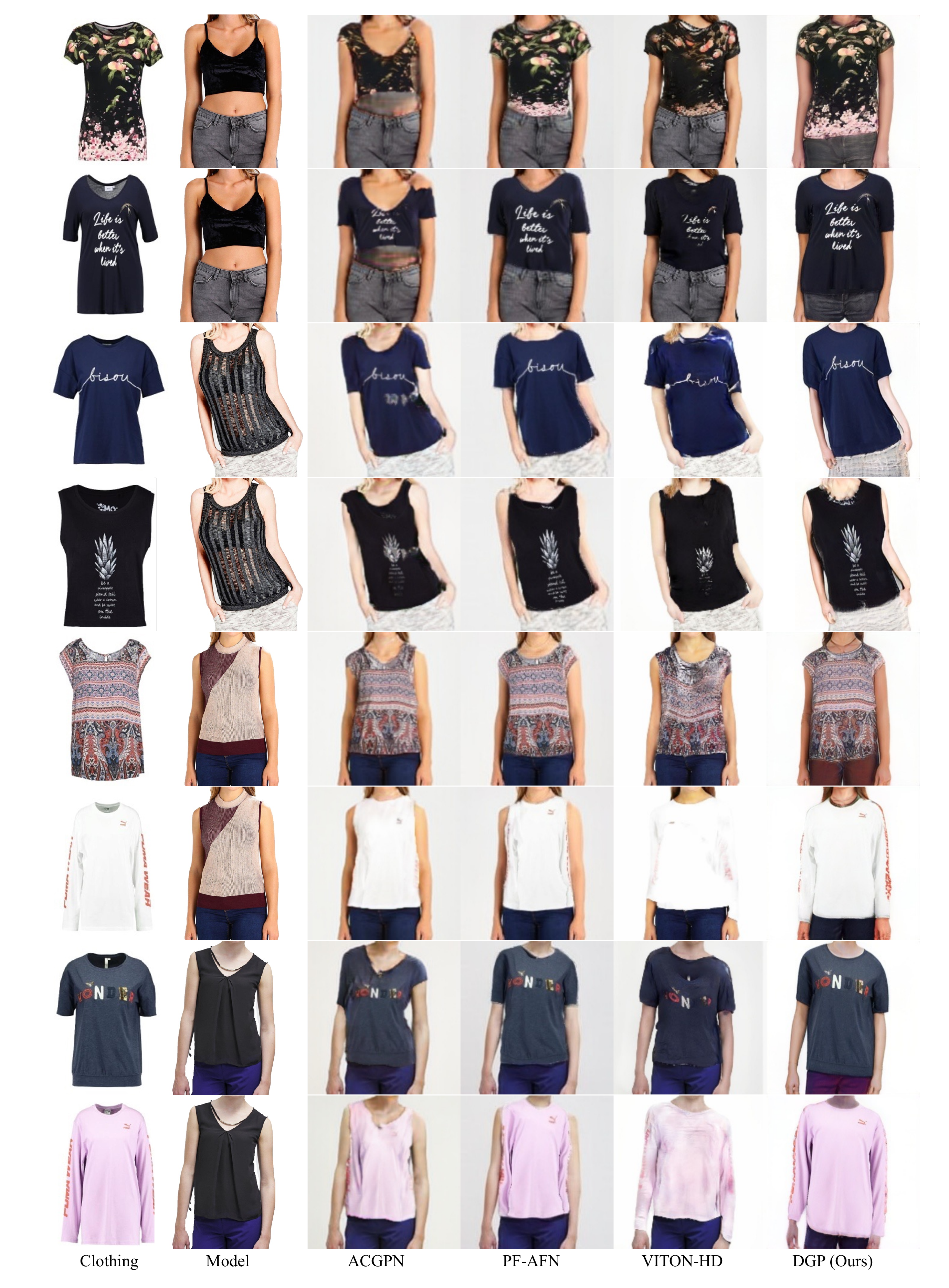}
  \caption{More visual results of qualitative comparison on the MPV dataset.}\label{fig: SM_compare_MPV}
\end{figure}

\clearpage

\section*{M Qualitative Comparison on CMI and MPV Dataset}
\addcontentsline{toc}{section}{M Qualitative Comparison on CMI and MPV Dataset}
More qualitative results on CMI and MPV datasets are reported of the proposed weakly supervised DGP method compared with the other three supervised competitors, VITON-HD, PF-AFN, and ACGPN. Please refer to Fig. \ref{fig: SM_compare_CMI} for more details on the CMI dataset and Fig. \ref{fig: SM_compare_MPV} on the MPV dataset.

\clearpage
\twocolumn
%%%%%%%%% REFERENCES
% {\small
% \bibliographystyle{ieee_fullname}
% \bibliography{egbib}
% }

\end{document}